\documentclass[11pt]{article}

\usepackage{tikz}
\usepackage{tkz-berge}
\usetikzlibrary{positioning}
\usetikzlibrary{arrows.meta}

\usepackage{geometry}
\usepackage{amsfonts}
\usepackage{amssymb}
\usepackage{amsmath}
\usepackage{bbm}
\usepackage{mathtools}
\usepackage{etex,etoolbox}
\usepackage{environ}
\usepackage{amsthm}
\usepackage{latexsym}
\usepackage{enumitem}
\usepackage{hyperref}
\usepackage{caption}
\hypersetup{
  colorlinks   = true, 
  urlcolor     = blue, 
  linkcolor    = blue, 
  citecolor   = blue
}

\usepackage{multicol}
\usepackage[ruled, vlined]{algorithm2e}
\makeatletter
\newtheorem*{rep@theorem}{\rep@title}
\newcommand{\newreptheorem}[2]{%
\newenvironment{rep#1}[1]{%
 \def\rep@title{#2 \ref{##1}}%
 \begin{rep@theorem}}%
 {\end{rep@theorem}}}
\makeatother

\newtheorem*{question}{Question}

\newreptheorem{theorem}{Theorem}

\newreptheorem{lemma}{Lemma}
\theoremstyle{definition}

\newtoks\magicAppendix
\magicAppendix={}
\newtoks\magictoks
\newif\iflater
\laterfalse
\long\def\later#1{\magictoks={#1}%
  \edef\magictodo{\noexpand\magicAppendix={\the\magicAppendix \par
    \noexpand\setcounter{theorem}{\arabic{theorem}}%
    \the\magictoks}}%
  \magictodo}
\long\def\both#1{\later{#1}\the\magictoks}
\def\magicappendix{\latertrue \the\magicAppendix}

\DeclareMathOperator{\perm}{perm}

\title{Formalizing the presumption of independence}
\author{Paul Christiano, Eric Neyman, Mark Xu \\ Alignment Research Center}

\begin{document}

\DeclarePairedDelimiter\parentheses{\lparen}{\rparen}
\DeclarePairedDelimiter\brackets{\lbrack}{\rbrack}
\DeclarePairedDelimiter\braces{\lbrace}{\rbrace}

\newcommand{\eps}{\varepsilon}
\newcommand{\of}[1]{\parentheses*{#1}}
\newcommand{\ofc}[2]{\parentheses*{#1\left\vert#2\right.}}
\newcommand{\setc}[2]{\left\{#1\left\vert#2\right.\right\}}
\renewcommand{\P}[1]{\mathbb{P}\of{#1}}
\newcommand{\Pc}[2]{\mathbb{P}\ofc{#1}{#2}}
\newcommand{\bof}[1]{\brackets*{#1}}
\newcommand{\brof}[1]{\braces*{#1}}
\newcommand{\Eof}[2]{\mathbb{E}_{#1}\bof{#2}}
\newcommand{\E}[1]{\mathbb{E}\bof{#1}}
\renewcommand{\k}[1]{\kappa\of{#1}}
\newcommand{\khat}[1]{\hat{\kappa}\of{#1}}
\newcommand{\kn}[1]{\hat{\kappa}^+\of{#1}}
\newcommand{\kp}[1]{\hat{\kappa}^-\of{#1}}
\newcommand{\khati}[1]{\hat{\kappa}_i\of{#1}}
\newcommand{\khatj}[1]{\hat{\kappa}_j\of{#1}}
\newcommand{\sof}[1]{\sigma^2\of{#1}}
\newcommand{\Cov}[1]{\operatorname{Cov}\of{#1}}
\newcommand{\cov}[1]{\operatorname{Cov}\of{#1}}
\newcommand{\Var}[1]{\operatorname{Var}\of{#1}}
\newcommand{\var}[1]{\operatorname{Var}\of{#1}}
\newcommand{\poly}[1]{\operatorname{poly}\of{#1}}
\newcommand{\set}[1]{\left\{#1\right\}}
\newcommand{\Pof}[2]{\mathbb{P}_{#1}\of{#2}}
\newcommand{\Pofc}[3]{\mathbb{P}_{#1}\of{#2\left\lvert#3\right.}}
\newcommand{\Pai}{\mathbb{P}_{\mathrm{AI}}}
\newcommand{\fai}{f_{\mathrm{AI}}}
\newcommand{\Ph}{\mathbb{P}_{\mathrm{H}}}
\newcommand{\fh}{f_{\mathrm{H}}}
\newcommand{\bits}[1]{\brof{0, 1}^{#1}}
\newcommand{\nbits}{\bits{n}}
\newcommand{\range}[1]{\set{1, 2, \ldots, #1}}
\newcommand{\nsat}{\# \mathrm{SAT}}
\newcommand{\nC}{\# \of{C}}
\newcommand{\sha}{\text{SHA-256}}
\newcommand{\sig}[3]{#1\colon #2 \rightarrow #3}
\newcommand{\pc}{\P{C}}
\newcommand{\ec}{\E{C}}
\newcommand{\loss}{\mathcal{L}}
\newcommand{\dist}{\mathcal{D}}
\newcommand{\cat}{\mathcal{C}}
\newcommand{\cattext}{\text{``catastrophe''}}
\newcommand{\abs}[1]{\left|#1\right|}
\newcommand{\nth}{n^{\text{th}}}
\newcommand{\fifth}{5^{\text{th}}}
\renewcommand{\phi}{\varphi}
\renewcommand{\epsilon}{\varepsilon}

\newcommand{\X}{\mathcal{X}}
\newcommand{\inner}[2]{\langle #1, #2 \rangle}

\date{}
\maketitle

\begin{abstract}
    Mathematical proof aims to deliver confident conclusions,
    but a very similar process of deduction can be
    used to make uncertain estimates
    that are open to revision.
    A key ingredient in such reasoning
    is the use of a ``default'' estimate of $\E{XY} = \E{X} \E{Y}$
    in the absence of any specific information about the correlation
    between $X$ and $Y$,
    which we call \emph{the presumption of independence}.
    Reasoning based on this heuristic is commonplace,
    intuitively compelling, and often quite successful---but completely
    informal.

    In this paper we introduce the concept of a heuristic estimator
    as a potential formalization of this type of defeasible reasoning.
    We introduce a set of intuitively desirable coherence properties
    for heuristic estimators that are not satisfied by any existing candidates.
    Then we present our main open problem:
    is there a heuristic estimator that formalizes intuitively
    valid applications of the presumption of independence
    without also accepting spurious arguments?
\end{abstract}

Many formally-specified questions are very hard to settle with proofs.
There are famous examples like the twin prime conjecture,
but also countless more mundane examples like how quickly the temperature
of a simulated room would change if the window were opened.

Even when we cannot prove a theorem, we can often deductively
arrive at a reasonable best guess
about the truth of a claim or the behavior of a system.
We can make probabilistic arguments about the structure of the primes
to estimate the density of twin primes,
or about small molecules moving randomly in order to estimate the rate of heat
transfer.

This reasoning requires making
best guesses about quantities that
we can't calculate exactly.
We can often do this using the \emph{presumption of independence}:
when trying to estimate $\E{XY}$ without any
knowledge about the relationship between $X$ and $Y$,
we can use $\E{X}\E{Y}$ as a default guess
rather than remaining completely agnostic.
For example, we can provisionally treat ``$x$ is prime''
and ``$x+2$ is prime'' as independent,
or treat the velocities of different air molecules
as uncorrelated.

This principle is sufficient to make
plausible estimates about a very wide range of mathematical quantities.
But it is not clear how to formalize this kind of defeasible
reasoning,
nor is it clear how to generalize our default guess
to the situation where we have arbitrary partial information
about how $X$ and $Y$ are related.

%
%

Heuristic reasoning using the presumption of independence
is distinct from running experiments or Monte Carlo simulations.
We are not merely observing a lot of twin primes
and inferring that there are probably infinitely many of them,
or running simulations of a room and observing how quickly the temperature
changes---we have found a good reason that our answer \emph{should be}
right unless there is additional structure
that we've overlooked which changes the answer.


We emphasize that this is not a novel proposal;
the presumption of independence is a common ingredient
in existing heuristic arguments
and has been explicitly articulated in essentially this form by \cite{tao}.
The purpose of this paper is to clarify the meta-problem of formalizing
this principle.

In Sections~\ref{twinprimes} and \ref{otherexamples}
we discuss informal examples of such reasoning
in number theory, combinatorics, and dynamical systems.
In Section~\ref{verifiers} we introduce
the concept of a \emph{heuristic estimator}
to formalize defeasible reasoning based on heuristic arguments.
In Section~\ref{desiderata} we introduce
a set of coherence conditions for heuristic estimators
which we believe should be satisfied by any adequate formalization
of the presumption of independence.
In Section~\ref{goal}
we precisely state the problem of finding a heuristic
estimator that formalizes a given set of
informal heuristic arguments.
Finally in Section~\ref{circuitsection} we propose
heuristic evaluation of boolean circuits
as a simple domain for studying heuristic estimators.

In the appendices we
discuss a number of subtleties and conjectures,
describe a simple formalization of the presumption
of independence that proves to be inadequate,
and discuss potential applications of heuristic arguments in machine learning.

\section[twin primes]{Example: the twin prime conjecture}\label{twinprimes}%

There are many existing examples of heuristic arguments,
especially in number theory;
\cite{tao} presents the twin prime conjecture as a simple example,
and in this section we essentially reiterate that presentation.

\begin{question}
    A twin prime pair is a pair of integers $\of{x, x+2}$ which are both prime.
    How many twin prime pairs are there with $x \leq N$?
\end{question}
By the prime number theorem,
a \emph{random} integer between $1$ and $N$
is prime with probability roughly\footnote{Throughout this section
we will ignore $o\of{\frac 1{\ln N}}$ correction terms.}
$\frac 1{\ln{N}}$.
So we have:
\begin{align*}
    \Pof{x \sim \range{N}}{\text{$x$ is prime}} = \Pof{x \sim \range{N}}{\text{$x+2$ is prime}} = \frac 1{\ln{N}}
\end{align*}
However, it is extraordinarily difficult
to calculate $\P{\text{$x$ is prime and $x + 2$ is prime}}$.
To make a best guess about this probability,
we will need to make some defeasible assumption:
\begin{description}
    \item[The presumption of independence.] If we have estimates for $\P{A}$ and $\P{B}$
        but know nothing about how $A$ and $B$ are related,
        then we presume that the events are independent and estimate $\P{A \wedge B} \approx \P{A} \P{B}$.
        This presumption can be overturned, and our estimate revised, if we later notice a way that $A$ and $B$ are related.

\end{description}
This principle is called the ``basic heuristic'' in \cite{tao},
and following their usage we will call an argument using it a
\emph{probabilistic heuristic argument}.
This principle seems almost inevitable
if we are committed to making
\emph{some} best guess about $\P{A \wedge B}$---after all we have
no reason to guess either a positive or negative correlation.

Using the presumption of independence,
we estimate:
\begin{align*}
    \Pof{x \leq N}{\text{$x$ is prime} \wedge \text{$x+2$ is prime}}
    &\approx \Pof{x \leq N}{\text{$x$ is prime}} \times \Pof{x \leq N}{\text{$x+2$ is prime}} \\
    &= \frac 1{\ln{N}} \times \frac 1{\ln N} = \frac 1{\ln^2 N}.
\end{align*}
So we expect $\frac {N}{\ln^2 N}$ twin primes less than $N$.
The twin prime conjecture is the statement that there are infinitely many twin primes;
by applying the presumption of independence again\footnote{
    The expected number of twin primes less than $N$
    approaches infinity as $N$ grows.
    So if we treat each event of the form ($x$ is prime and $x+2$ is prime)
    as independent,
    then with probability $1$ infinitely many of them occur.
}
we estimate $\P{\text{twin prime conjecture}} = 1$.

Our estimate for the number of twin primes is uncertain for two reasons:
\begin{description}
    \item[Chance.]
        There may be surprisingly few or surprisingly
        many twin primes ``by chance.''
        For example, this same methodology expects that a random pair $(x, x+2)$
        between $10,000$ and $10,100$ has about a $1/100$ chance of being a twin prime,
        and so on average there will be about $1$ twin prime pair in that range.
        But we would not be too surprised to find that there were actually no twin primes in the interval,
        or that there were multiple.\footnote{If we apply the presumption
            of independence again then we can predict that the number
            of twin prime pairs in the interval is approximately Poisson with
            mean 1---the same as the count of heads
            if you flip $100$ coins each with a $1\%$ probability of heads.}
    \item[Defeasibility.]
        More importantly, this estimate could change completely if we later noticed a reason that $\of{\text{$x$ is prime}}$
        and $\of{\text{$x+2$ is prime}}$ are correlated.

        For example, if we had instead been trying to estimate the number of pairs $\of{x, x+1}$
        that are both primes,
        we would have \emph{also} concluded that there should be about $\frac N{\ln^2 N}$
        and that there should be infinitely many with probability $1$.
        But eventually we may notice that at least one of $x$ and $x+1$ is divisible by $2$,
        and so for $x > 2$ these events are perfectly anticorrelated.
        So our conclusion was wrong even though we gave it probability of $1$.
        The probabilities we assign do not capture the possibility of this kind of revision---they
        quantify only the uncertainty from ``chance'' and not from defeasibility.

        For the twin prime conjecture there are a few considerations that slightly change the estimate $\frac {N}{\ln^2 N}$.
        Most importantly, if $x$ is prime
        then $x+2$ is also odd and hence twice as likely to be a prime.
        The net effect of all known corrections is to increase
        our estimate by about 30\% from
        $\frac {N}{\ln^2 N}$ to $\frac {2 C_2 N}{\ln^2 N}$,
        where $C_2 = 0.660\ldots$ is called the \emph{twin prime constant},\footnote{
            This constant is derived from
            the obvious negative correlation between the events
            ($p$ divides $x$) and ($p$ divides $x+2$) for $p > 2$.
            The Hardy-Littlewood conjecture implies
            that this is the true asymptotic density of the twin primes,
            i.e. that there are no further corrections.
            This conjecture appears to agree with experimental data
            but is expected to be extremely
            difficult to prove.
            There are other correction terms, which meaningfully change the expected number
            of twin primes between $10,000$ and $10,100$
            but are asymptotically negligible in $N$.
        }
        but until we have a proof we cannot rule
        out the possibility of finding a new consideration
        that totally changes our estimate.
\end{description}

Despite these limitations,
we think that probabilistic heuristic arguments
can give us reasonable best guesses about the truth of
mathematical statements.

The other side of these limitations is that it is typically \emph{much}
easier to make a heuristic estimate than to find a proof.
Intuitively, a heuristic estimate represents a best guess given whatever
structure and correlations we have noticed so far,
whereas a proof requires ruling out the possibility of \emph{any} other correlations
or coincidences.
This is much harder and usually requires completely different techniques.

\section{Other examples}\label{otherexamples}

We can use the presumption of independence to produce heuristic estimates
across a wide variety of domains:
\begin{description}
    \item[Diffusion.] Suppose that I have a frictionless pool table
        with a line down the middle dividing it in half.
        I place 15 perfectly elastic pool balls at random on the left half of the table each with
        an initial velocity of 1 meter per second in a random direction.
        After twenty seconds, what is the probability that most of the balls are still
        on the left half of the table?

        Exactly tracking how the distribution of balls changes over time is completely intractable.
        But we could summarize it by separately considering the distribution
        over each ball's position and velocity.
        If we treat these quantities as independent for different balls,
        then it becomes easy to track how they evolve over time.
        Under this simplification the positions quickly converge to uniform.
        Within 20 seconds each ball has almost exactly a $50\%$ chance of being on either half of the table,
        and so the probability of most of them being on the left half is also $50\%$.
        (We discuss this example in more detail in Appendix~\ref{ballexample}.)
    \item[Hash functions.] SHA-256 is a complex circuit with $256$ bit outputs.
        What is the probability that there exists a $256$ bit string $x$ such that $\sha\of{x}$
        is all zeros?

        To answer this question we want to understand the output distribution of SHA-256
        if we sample the input bits uniformly and independently.
        This is very hard to compute exactly,
        but it is quite easy to compute the probability distribution over each intermediate
        value computed by SHA-256 \emph{if we assume that each operation's inputs are independent}.
        Under this approximation we find that essentially every intermediate value is uniformly random,
        and in particular the output bits are unbiased.

        If we further assume that those output bits are independent,
        then there is a $2^{-256}$ chance that any given value $\sha\of{x}$
        has all $256$ bits equal to $0$.
        If these different values of $\sha\of{x}$ are themselves independent,
        then there is a probability of $1 - \of{1 - 2^{-256}}^{2^{256}} \approx 1 - 1/e$
        that at least one output is all zeros.
    \item[The prime number theorem.] Our analysis of the twin
        prime conjecture relied heavily on
        the claim that a random number $x$ has a $\frac 1{\ln x}$ chance of being prime.

        We can derive this fact heuristically by noticing that $x$ is prime if and only if
        it has no prime divisors, and treating each event $p|x$
        as independent with probability $\frac 1p$.
        This implies \[\P{\text{$x$ is prime}} = \prod_{\text{prime $p$} < x} \of{1 - \frac 1p}\]
        and gives us an estimate for $\P{\text{$x$ is prime}}$ that depends
        on the number and distribution of smaller primes.
        By solving the resulting recurrence relation we conclude that
        $\P{\text{$x$ is prime}} = \frac 1{\ln x} + O\of{\frac 1{\ln^2 x}}$.
\end{description}

Note that in all of these cases the \emph{only} heuristic step
is the presumption of independence---the rest of the argument is deductively
valid.
We walk through more examples in Appendix~\ref{examples},
each of which is also a deductively valid argument
combined with a suitable generalization of the presumption of independence.

This is not the only possible kind of heuristic argument.
For example, we might conclude that a theorem is likely to be true
based on checking enough special cases,
or conclude that a theorem is likely to be false
because it involves a constant like $3.14158$ that looks like
it \emph{should} be $\pi$.

But the presumption of independence seems like an extremely
general and powerful tool,
which is sufficient to produce useful
heuristic estimates across a broad range of domains.
This is easiest to assess in mathematics and especially number theory,
where we believe there are probabilistic heuristic arguments
for a significant majority of open problems,\footnote{
    For example, we reviewed the list of 105 pages in the Wikipedia
    category \href{https://en.wikipedia.org/wiki/Category:Unsolved_problems_in_number_theory}{``Unsolved problems in number theory.''}
    Based on random sampling, we estimate that for
    more than 75\% of these conjectures
    the authors would be able to find a probabilistic heuristic argument
    that we find convincing.
    (About 30\% are justified by the Cram\'er random model of the primes,
    and about 6\% are justified by the kind of Diophantine equation heuristic
    discussed in Appendix~\ref{flt}.)
    The counterexamples primarily involve non-elementary statements or arguments
    that are difficult to assess without expertise in number theory,
    and we believe that a domain expert could probably give probabilistic
    heuristic arguments for more than 90\% of these statements.

    Those estimates should not be taken too seriously, especially given that
    we don't have a formalization of heuristic arguments
    that we can use to reduce experimenter bias
    or assess how often it is possible to give spurious arguments
    for incorrect conclusions.
    But we think they still give some general indication that the presumption
    of independence is often sufficient to justify plausible conjectures.
%
}
but we believe that it is also effective
in other domains where efficacy is harder to quantify.
%
\section{Heuristic estimators}\label{verifiers}

What would it look like to formalize this kind of reasoning?

We can formalize a traditional proof system
by specifying a language for proofs
and defining a proof verifier $V$:
an efficient program which takes as input a statement $\phi$
and a putative proof $\pi$, and then outputs a judgment $V\of{\phi, \pi} \in \set{\top, \bot, ?}$.
The outputs $\top$ or $\bot$ indicate that $\pi$ was a proof or disproof of $\phi$
and in these cases we might say that $V$ confidently ``believes'' $\phi$ to be true or false.
The output $?$ indicates that $\pi$
was not a valid proof and so $V$ is agnostic about $\phi$.

\newcommand{\tP}{\widetilde{\mathbb{P}}}

We will aim to formalize heuristic arguments by specifying
a language for heuristic arguments
and defining an analogous \emph{heuristic estimator} $\widetilde{\mathbb{P}}$:
an efficient program which takes as input a statement $\phi$
and a \textbf{set} of heuristic arguments $\pi_1, \pi_2, \ldots, \pi_n$,
then outputs a best guess $\tP\of{\phi, \pi_1, \ldots, \pi_n} \in [0, 1]$ about the 
probability of $\phi$.

The major conceptual difference between a heuristic estimator
and a proof verifier
is that a heuristic estimator \emph{always} outputs a best guess in light
of the available arguments,
whereas a proof verifier effectively remains agnostic until finding a proof.
These estimates are subject to revision and need not be calibrated,
but we do still expect them to satisfy simple coherence properties
(see Section~\ref{desiderata}).
As a special case, $\tP\of{\phi}$ should produce a default estimate
before seeing any arguments at all.\footnote{
    For example, we could define a very bad estimate $\tP\of{\phi}$ based purely
    on the presumption of independence and the structure of $\phi$.
    We can take
    $\tP\of{A \wedge B} = \tP\of{A} \tP\of{B}$,
    and treat $\tP\of{\forall x \colon \phi\of{x}}$
    as an a very large conjunction.
    As a result, almost any universally quantified statement will have $0$
    probability by default.
}

The reason we consider a set of arguments rather than just one
is that any given argument is defeasible and open to revision.
If Alice points out a reason to think that $\phi$ is true
and Bob points out a reason to think that $\phi$ is false,
we want to be able to combine those arguments to arrive
at an all-things-considered best guess about $\phi$.
This was not necessary for proof verifiers
because a single proof settles the question.

Our goal is to find a natural heuristic estimator $\tP$
that is able to recognize the kind of argument presented in Section~\ref{twinprimes}.
That is, after seeing such an argument it should output
that the twin prime conjecture is almost certainly true,
and then it should only revise that conclusion if given another argument
$\pi_i$ that undermines one of the independence
assumptions and suggests an alternative estimate.
We formalize this goal in Section~\ref{goal}.

\newcommand{\tE}{\widetilde{\mathbb{E}}}
\newcommand{\V}[1]{\tE\of{#1}}
\newcommand{\Vpi}[1]{\V{#1, \pi_1, \ldots, \pi_n}}

Rather than only evaluating the truth of propositions,
we will generalize further to heuristic estimators $\tE$ for arbitrary quantities.
In this case we take $X$ to be a formal expression defining a real number,
and interpret $\Vpi{X}$ as a ``subjective expected value'' of $X$.\footnote{
    This definition is most straightforward if $X$ is bounded,
    i.e. if we have a proof that $\ell \leq X \leq h$ for some particular real numbers $\ell$ and $h$.
    If there are no provable bounds on $X$ then the expectation may be infinite or undefined.
    For now we will set this issue aside;
    a concerned reader can restrict their attention to quantities $X \in [0, 1]$.
}
Of course we can recover $\tP$ as the expectation of the indicator
function $\mathbbm{1}_{\phi}$.

\subsection{A bad example of a heuristic estimator}

To illustrate the definition,
we can define a heuristic estimator $\tE$
that treats $X$ as uniformly random between the lowest
and highest possible value:
\begin{itemize}
    \item \textbf{What is an argument $\pi_i$?} An argument $\pi_i$
        must be a proof
        that $\ell \leq X \leq h$ for some real numbers $\ell$ and $h$.
    \item \textbf{What is $\Vpi{X}$?}
        Let $\ell^*$ be the maximum of the lower bounds proven by any of the $\pi_i$,
        and let $h^*$ be the minimum of the upper  bounds.
        Define $\Vpi{X}$
        to be the average of those bounds $\frac{\ell^* + h^*}{2}$,
        with the convention that $\frac{-\infty + \infty}{2} = 0$
        so that $\V{X} = 0$.
\end{itemize}
We consider this heuristic estimator extremely unreasonable.
To see why, suppose that $A, B \in \set{0, 1}$
have complex definitions such that it is hard to prove anything about them
or about how they relate.
We would expect a good heuristic estimator to treat each of them as uniformly random,
and to converge to an estimate $\Vpi{AB} = \frac 14$
once all relevant arguments are pointed out.
But if the only thing we can \textbf{prove} is that $AB \in \set{0, 1}$,
then this estimator will instead converge to the estimate $\Vpi{AB} = \frac 12$.

In fact, after seeing the relevant arguments this estimator converges to:
\begin{gather*}
    \V{AB} = \V{A(1-B)} = \V{(1-A)B} = \V{(1-A)(1-B)} = \frac 12 \\
    \V{AB  + A(1-B) + (1-A) B + (1-A)(1-B)} = \V{1} = 1
\end{gather*}
and so $\tE$ is not even linear.

%

\section{Desirable properties for heuristic estimators}\label{desiderata}

A heuristic estimator $\tE$ should behave like an expectation.
That is, for any sequence of arguments $\pi_1, \ldots, \pi_n$ it should satisfy:
\begin{itemize}
    \item \textbf{Constant expectations.} For any constant $c$,
        \[\Vpi{c} = c.\]
    \item \textbf{Linearity of expectation.}
        For any quantities $X, Y$ and constants $a, b$,
        \[\Vpi{aX + bY} = a\Vpi{X} + b\Vpi{Y}.\]
\end{itemize}
A good estimator $\tE$ should revise
its estimates based on arguments,
which should be at least as expressive
as traditional proofs:
\begin{itemize}
    \item \textbf{Respect for proofs.}
        If $\pi$ is a proof
        that $X \geq 0$,
        then there should be an analogous heuristic argument $\pi_*$
        such that for any $\pi_1, \ldots, \pi_n$,
        \[ \Vpi{X, \pi_*} \geq 0.\]
\end{itemize}
This
property depends on the choice of proof system;
we are looking for heuristic estimators that respect
as many proofs as possible.
Together with linearity of expectation,
respect for proofs implies that if $X$ and $Y$
are provably equal,
then there is a $\pi_*$ such that $\Vpi{X, \pi_*} = \Vpi{Y, \pi_*}$
for any $\pi_1, \ldots, \pi_n$.

A reasonable estimator $\tE$ should not
revise its beliefs if we provide an irrelevant argument $\pi_*$,
or if we repeat or rearrange arguments:
\begin{itemize}
    \item \textbf{Independence of irrelevant arguments (informal).}
        If $\pi_*$ is irrelevant to the value of $X$,
        then
        \[\Vpi{X, \pi_*} = \Vpi{X}.\]
    \item \textbf{Invariance to repetition and rearrangement.}
        If $\set{\pi_1, \ldots, \pi_n} = \set{\pi_1', \ldots, \pi_m'}$,
        i.e. if the two sequences of arguments are the same up to repetition and rearrangement, then
        \[\V{X, \pi_1, \ldots, \pi_n} = \V{X, \pi_1', \ldots, \pi_m'}.\]
\end{itemize}

Finally, we are particularly interested in heuristic
estimators that capture the presumption of independence.
\begin{itemize}
    \item \textbf{Presumption of independence (informal).}
        If $\pi_1, \ldots, \pi_n$
        do not provide any reason to think that $X$ and $Y$ are related,
        then
        \[\Vpi{XY} = \Vpi{X} \Vpi{Y}.\]
\end{itemize}
These six properties are not necessarily sufficient
to conclude that a heuristic estimator is reasonable,
but we are not aware of any estimator that satisfies them.
We believe that finding such an estimator would be
a promising step forward.

\subsection{Heuristic arguments sometimes make estimates worse}\label{cherrypickshort}

One desirable property was conspicuously missing from the above list:
\begin{itemize}
    \item \textbf{Monotonic improvement.}
        For any $\pi_*$ and any $X$,
        \[\abs{\V{X, \pi_*, \pi_1, \ldots, \pi_n} - X} \leq \abs{\V{X, \pi_1, \ldots, \pi_n} - X}.\]
\end{itemize}

Unfortunately, no matter how good an estimator $\tE$ we find,
we do not expect monotonic improvement.
That is, we think it is possible for valid arguments
to push even an ideal reasoner's beliefs
in the wrong direction.

To see this, suppose we are trying to estimate
$A+B+C$ where $A, B, C \in \set{+1, -1}$.
Assume that $\V{A} = \V{B} = \V{C} = 0$,
so $\V{A+B+C} = 0$.
Suppose that $\pi_A$ is a proof that $A = 1$.
Then we expect $\V{A+B+C, \pi_A} = 1$.
But it may turn out by chance that $B=C=-1$,
in which case $A+B+C=-1$ and the argument $\pi_A$
happened to push $\tE$'s estimate in the wrong direction.
This means that even if we are searching for arguments
in an unbiased way,
they will sometimes happen to make our estimate worse
by chance.
And if someone searches for adversarially misleading
estimates,
they will usually be able to succeed.

%
%

In Appendix~\ref{cherrypicking} we discuss a sequence
of increasingly severe versions of this problem,
and explore the behavior of heuristic estimators when given adversarially-selected
arguments. 
Despite the fact that arguments do not always improve estimates,
we still believe that formalizing heuristic arguments
can help clarify which arguments we ought to consider valid
and how we should update our beliefs in light of them.

\section{Formalizing intuitively valid heuristic arguments}\label{goal}

\newcommand{\wtp}{\widetilde{\pi}}

One of our main goals is to find a heuristic estimator $\tE$ that
is able to accept as many intuitively valid heuristic arguments
as possible
without also accepting spurious arguments.
In this section we try to make this goal more precise.

We have already seen a few examples of informal heuristic arguments
based on the presumption of independence.
In Appendix~\ref{examples} we present three more detailed examples.
Each example can be described as a triple $\of{X, \mu, \wtp}$,
where $\wtp$ is an informal heuristic argument that $\E{X} = \mu$.
For example,
$X$ could be the number of twin primes less than $2^{256}$
and $\wtp$ could be the informal argument in Section~\ref{twinprimes}.

For a given triple $\of{X, \mu, \wtp}$,
we can capture whether $\tE$ accepts $\wtp$
by asking whether there exists a
formalization $\pi$ of $\wtp$ such that 
\[\V{X, \pi} = \mu.\]
It is less clear how to precisely state the requirement that $\tE$ does not also accept
spurious arguments because we have not defined what a ``spurious argument'' is.

Fortunately, in many cases we would be very surprised to find significant
revisions to the estimate $\mu$.
For example,
any significant revision to the heuristic estimate for the number of twin primes in
Section~\ref{twinprimes} would be a major and surprising development
in number theory.
In these cases,
we think that \emph{any} argument that changes $\tE$'s estimate from $\mu$ is likely to be spurious.
So we expect $\tE$ to satisfy:
\begin{equation}\label{goaleq}
\exists \pi \colon \forall \set{\pi_1, \pi_2, \ldots, \pi_n} \ni \pi \colon
\Vpi{X} \approx \mu,
\end{equation}
where it should be straightforward (but potentially laborious)
to construct $\pi$ from $\wtp$.
In words, it should be possible to produce a formalization $\pi$ of $\wtp$
such that if we present $\pi$ to $\tE$ it produces an estimate close\footnote{
    The quantitative closeness depends on the problem,
    and in particular on how much we think that further valid arguments
    should be able to change $\tE$'s views.
    For example, in the case of estimating the number of twin primes less than $N$,
    we expect the correction to be asymptotically negligible in $N$,
    and any non-negligible correction would contradict
    the Hardy-Littlewood conjecture.
    In the case of estimating the probability of a zero of SHA-256,
    it is easy to find arguments resulting in adjustments on the order of $2^{-256}$,
    but any argument leading to a revision of say $2^{-128}$ would be a major development
    in cryptanalysis.
} to $\mu$,
even if we also provide $\tE$ a set of adversarially misleading arguments.

So any set of triples $\of{X, \mu, \wtp}$ leads to a simple open problem:
find a heuristic estimator that satisfies Equation~\ref{goaleq}
for as many triples in that set as possible.

Of course it is possible to satisfy this property for any
finite set of triples $(X, \mu, \wtp)$
by specifying the expected answers directly as part of the definition of $\tE$.
So to make the problem challenging we want to search for an $\tE$ that also works for
a larger set of similar ``held out'' examples $\tE$.\footnote{
    Alternatively we could search for a sufficiently simple estimator that
    satisfies Equation~\ref{goaleq}.
    Or we could informally evaluate a proposed estimator
    based on an intuitive judgment about whether it looks like it would it generalize
    to new claims.
} Fortunately it is easy to generate a very large number of
examples of intuitively compelling heuristic arguments for which
significant revisions would be surprising,
leading to a large set of triples $\of{X, \mu, \wtp}$
that can be used to evaluate a proposed estimator $\tE$.
In this document we provide only a small list of examples
to illustrate the problem,
but we expect to publish a larger list of examples in the future
and to maintain a large private
``test set'' that we can use to evaluate proposed estimators.

The wider the distribution for which $\tE$ works the better,
but finding an estimator $\tE$ for even a narrow domain already seems challenging.
For example,
we believe that a significant majority of plausible conjectures in number theory
are supported by a probabilistic heuristic argument.
Some of those conjectures
can be settled by the Cram\"{e}r model of the primes
or the Diophantine equation heuristic described in Appendix~\ref{flt}.
But many of them require \emph{ad hoc} heuristic arguments,
and we think that it is a difficult challenge to write down a verifier
that satisfies Equation~\ref{goaleq} for a significant fraction of those cases.
While there are many simple ways to formalize more general probabilistic heuristic arguments,
most of them require unformalized judgment calls,
and we believe that any existing fully precise $\tE$ would also accept
spurious arguments for incorrect conclusions.

\section{Circuits as a setting to study heuristic arguments}\label{circuitsection}

We are ultimately interested in formalizing the entire
range of heuristic arguments that are used in mathematical practice.
But it is helpful to have a simplified setting
both to illustrate the challenge and to study candidate algorithms.

We propose circuit evaluation as a simple but challenging domain:
given a circuit $\sig{C}{\bits{n}}{\bits{}}$
estimate the probability that $C(z) = 1$ for a uniformly random input $z$.
In this section we describe the task,
present a very simple algorithm,
and discuss why we consider the challenge interesting.

\subsection{Task definition}\label{circuitdefinition}

\begin{figure}
    \centering
    \begin{tikzpicture}[
        roundnode/.style={
            circle, draw=black!60, fill=green!5, very thick, minimum size=7mm
        },
        naked/.style={
        },
        ]
        \node[naked] (z1) [label={[text=blue]15:$1$}] {$z_2$};
        \node[naked] (one) [left=of z1, label={[text=blue]15:$1$}] {$z_1$};
        \node[naked] (z2) [right=of z1, label={[text=blue]15:$0$}] {$z_3$};
        \node[roundnode] (onez1) [below=of one, minimum size=1.1cm, label={[text=blue]15:$1$}] {\footnotesize{OR}};
        \node[roundnode] (onez2) [below=of z1, label={[text=blue]15:$1$}] {\footnotesize{XOR}};
        \node[roundnode] (z1z2) [below=of z2, minimum size=1.1cm, label={[text=blue]15:$1$}] {\footnotesize{OR}};
        \node[roundnode] (A) [below left=of onez2, xshift=0.8cm, label={[text=blue]15:$1$}] {\footnotesize{AND}};
        \node[roundnode] (B) [below right=of onez2, xshift=-0.8cm, label={[text=blue]15:$1$}] {\footnotesize{AND}};
        \path[draw=none] (A) -- node (AB) {} (B);
        \node[roundnode] (C) [below =of AB, label={[text=blue]15:$0$}] {\footnotesize{XOR}};
        \draw[->] (one) -- (onez1);
        \draw[->] (z1) -- (onez1);
        \draw[->] (z1) -- (z1z2);
        \draw[->] (z2) -- (z1z2);
        \draw[->] (z2) -- (onez2);
        \draw[->] (one) -- (onez2);
        \draw[->] (onez1) -- (A);
        \draw[->] (onez2) -- (A);
        \draw[->] (onez2) -- (B);
        \draw[->] (z1z2) -- (B);
        \draw[->] (B) -- (C);
        \draw[->] (A) -- (C);
    \end{tikzpicture}
    \caption{A simple boolean circuit. The output node $x_m$
        is at the bottom of the figure,
        while the input nodes at the top of the figure
        are labeled with $z_1$, $z_2$, and $z_3$.
        In blue we have shown the result of evaluating the circuit
        on the input triple $\of{z_1, z_2, z_3} = \of{1, 1, 0}$.
        For example, the output XOR gate is equal to $0$
        because its two inputs are both equal to $1$
        and $\text{XOR}(1, 1) = 0$.
        So $C\of{1, 1, 0} = 0$.
    }\label{circuitexample}
\end{figure}
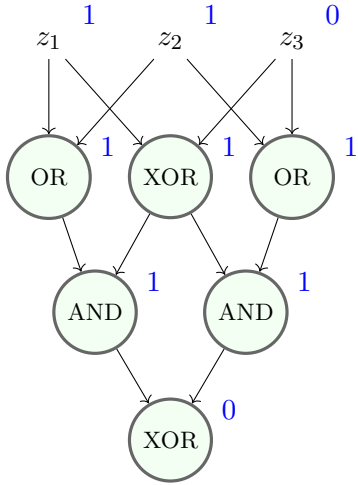

Informally, a boolean circuit is a recipe for computing an output value
$x_m$ by starting with a set of inputs $z_1, \ldots, z_n$
and then applying a fixed sequence of boolean operations.

Formally, a boolean circuit with $n$ inputs is defined as a set of $m$ nodes
$x_1, \ldots, x_m$,
where each node $x_k$ is either:
\begin{itemize}
    \item An input node labeled with an integer $i_k \in \set{1, \ldots, n}$.
    \item A binary gate labeled with a boolean operation
        $f_k \in \set{\text{AND}, \text{OR}, \text{XOR}, \ldots}$
        and the indices of two inputs $a_k, b_k \in \set{1, \ldots, k-1}$.
\end{itemize}
A simple circuit is depicted in Figure~\ref{circuitexample}.

To evaluate a circuit on an input $z = \of{z_1, \ldots, z_n} \in \bits{n}$,
we proceed through the nodes in order:
the value of an input node labeled with $i_k$ is equal to $z_{i_k}$,
and the value of a binary gate labeled with $f_k$ is equal to $f_k$
applied to the values of the two inputs $x_{a_k}$ and $x_{b_k}$.
The output $C\of{z}$
of the circuit is the value of the final node $x_m$.

We write $\pc$ for the probability that $C\of{z} = 1$ when $z$ is uniformly random.

We are interested in finding a heuristic estimator
$\Vpi{\pc}$
that satisfies the kind of desiderata introduced in Section~\ref{desiderata}
and is able to formalize a variety of intuitively valid heuristic arguments
about $\pc$ in the sense introduced in Section~\ref{goal}.
We will write $\V{C}$ instead of $\V{\pc}$.\footnote{
    This notational difference suggests a more subtle difference in how $\tE$
    actually performs the estimate.
    We will often describe heuristic estimators that
    effectively consider each input $z_i$
    as an unknown boolean variable with probability $1/2$,
    rather than estimators that consider a sum over the set of all possible inputs $z_i$.
    In particular, $\tE$ estimates $\pc$ in the same
    way that it would estimate the value of $C$ when run
    on a set of $m$ uncomputable and apparently unbiased inputs.
    For estimators with this form,
    it is more correct to talk about $\V{C(z)}$
    rather than $\V{\pc}$,
    where $z$ is a special symbol representing an unknown set of inputs
    specified to have a uniform distribution.
    These two perspectives are essentially equivalent due to linearity
    of expectation.
}

For example, if $\pi_1$ proves that the output of $C$ is equal
to the conjunction of $k$ unbiased and apparently unrelated intermediate values,
then we should have $\V{C, \pi_1} \approx 2^{-k}$.
If $\pi_2$ proves that actually two of these
intermediate values are almost always equal,
then that should cause $\V{C, \pi_1, \pi_2}$ to rise to roughly $2^{-k+1}$.
As we consider more and more intuitively compelling
arguments $\tE$ should continue to update in the expected way.

Instead of considering a heuristic estimator,
we could compute a Monte Carlo estimate for $\pc$
by randomly sampling inputs and calculating the empirical mean of $C(z)$.
A heuristic estimator can
have two advantages over the Monte Carlo estimator:
\begin{itemize}
    \item If $\pc$ is very close to $0$,
        then $\tE$ can be \emph{much} faster.
        It would require about $2^{256}$ samples
        to distinguish $\pc = 2^{-256}$ from $\pc = 2^{-512}$,
        but for many circuits we can make heuristic arguments
        that distinguish these cases using exponentially less time.
        This is similar to the use of propositional logic to establish
        a tautology without needing to consider every setting of every variable.
    \item We are interested in estimators $\tE$ that deterministically analyze
        the structure of $C$ rather than measuring $\pc$ by random sampling,
        because we think that this kind of analysis reveals something about \emph{why}
        $\pc$ takes on the value that it does.
        Although we cannot formalize this distinction precisely,
        we think it is important and discuss it in Appendix~\ref{explaining}.
\end{itemize}

\begin{figure}
    \centering
    \begin{tikzpicture}[
        roundnode/.style={
            circle, draw=black!60, fill=green!5, very thick, minimum size=7mm
        },
        naked/.style={
        },
        ]
        \node[naked] (z1) [label={[text=red]15:$\frac 12$}] {$z_2$};
        \node[naked] (one) [left=of z1, label={[text=red]15:$\frac 12$}] {$z_1$};
        \node[naked] (z2) [right=of z1, label={[text=red]15:$\frac 12$}] {$z_3$};
        \node[roundnode] (onez1) [below=of one, minimum size=1.1cm, label={[text=red]15:$\frac 34$}] {\footnotesize{OR}};
        \node[roundnode] (onez2) [below=of z1, label={[text=red]15:$\frac 12$}] {\footnotesize{XOR}};
        \node[roundnode] (z1z2) [below=of z2, minimum size=1.1cm, label={[text=red]15:$\frac 34$}] {\footnotesize{OR}};
        \node[roundnode] (A) [below left=of onez2, xshift=0.8cm, label={[text=red]15:$\frac 38$}] {\footnotesize{AND}};
        \node[roundnode] (B) [below right=of onez2, xshift=-0.8cm, label={[text=red]15:$\frac 38$}] {\footnotesize{AND}};
        \path[draw=none] (A) -- node (AB) {} (B);
        \node[roundnode] (C) [below =of AB, label={[text=red]15:$\frac {30}{64}$}] {\footnotesize{XOR}};
        \draw[->] (one) -- (onez1);
        \draw[->] (z1) -- (onez1);
        \draw[->] (z1) -- (z1z2);
        \draw[->] (z2) -- (z1z2);
        \draw[->] (z2) -- (onez2);
        \draw[->] (one) -- (onez2);
        \draw[->] (onez1) -- (A);
        \draw[->] (onez2) -- (A);
        \draw[->] (onez2) -- (B);
        \draw[->] (z1z2) -- (B);
        \draw[->] (B) -- (C);
        \draw[->] (A) -- (C);
    \end{tikzpicture}
    \caption{In red we have written the intermediate values $\V{x_k}$
        computed by assuming that all gates are independent
        for the simple circuit from Figure~\ref{circuitexample}..
        We obtain the estimate $\V{C} = 30/64 = \of{3/8}\of{5/8} + \of{5/8}\of{3/8}$.
        The estimate of $\V{C}$ would be correct if the two AND gates were independent
        and each equal to $1$ with probability $3/8$,
        but actually they are highly correlated (since they have a common input).
        The true value is $\pc = 2/8$: the circuit returns $1$
        if and only if $\of{z_1, z_2, z_3}$ is either $\of{1, 0, 0}$ or $\of{0, 0, 1}$.
    }\label{meanpropexample}
\end{figure}
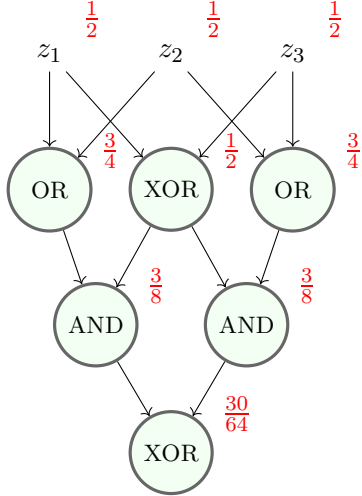

\subsection{A simple algorithm: assume all nodes are independent}\label{meanprop}

One of the simplest possible algorithms is to apply
the presumption of independence to every gate
in order to estimate $\V{x_k}$, the probability that node $x_k$
has value $1$ for random inputs $z_1, \ldots, z_n$:
\begin{itemize}
    \item If $x_k$ is an input node, then $\V{x_k} = \frac 12$.
    \item If $x_k = \text{AND}\of{x_{a_k}, x_{b_k}}$,
        then $\V{x_k} = \V{x_{a_k}} \V{x_{b_k}}$;
        if $x_k = \text{OR}\of{x_{a_k}, x_{b_k}}$
        then $\V{x_k} = 1 - \of{ 1 - \V{x_{a_k}} }\of{1 - \V{x_{b_k}}}$;
        and similarly for other functions.
\end{itemize}
Finally we output $\V{C} = \V{x_m}$.
We work through an example of this algorithm in Figure~\ref{meanpropexample}.

We can define more accurate estimates by tracking not only
the probabilities $\V{x_k}$ that individual node are $1$,
but various higher-order correlations amongst the nodes.
For example,
we might track the joint distribution of every pair of nodes $\V{x_i \wedge x_j}$,
or we might track the expectation of particular
large parities $\V{x_{i_1} \oplus x_{i_2} \oplus \cdots \oplus x_{i_k}}$
that are important for understanding the behavior of the circuit.

In Appendix~\ref{cumulantappendix} we present
an estimator $\tE$ which takes arbitrary ``advice'' $\pi_1, \ldots, \pi_k$
about which correlations to track,
and uses it to produce a heuristic estimate for $\pc$.
Unfortunately, this estimator often produces
implausible values $\V{C} \not\in [0, 1]$.
We are interested in
a better estimator
that is able to capture the same intuitively valid arguments about $\pc$
while also satisfying the desiderata from Section~\ref{desiderata}.


\subsection{Why care about circuits?}

We view heuristic circuit evaluation as a natural
generalization of verifying propositional tautologies.\footnote{
    To be more precisely analogous we could consider heuristic evaluation of
    \emph{formulas} instead of circuits,
    i.e. we could require that each node be used at most once
    as the input to another gate.
    This even simpler problem also seems challenging
    and interesting.
}
Rather than asking whether an expression is \emph{guaranteed} to be true or false
without knowing anything about its inputs,
we are instead asking how \emph{likely} it is to be true
given uniform ignorance about its inputs.

We are particularly optimistic about formalizing
informal heuristic arguments that do not involve quantifiers or abstractions.
Such heuristic arguments seem to be analogous to proofs in propositional
logic, but despite their simplicity
we nevertheless cannot write down any estimator $\tE$ which is able to capture them.

We believe that heuristically evaluating circuits is a stepping stone
to formalizing general heuristic arguments in the same
way that propositional logic is a stepping stone towards first-order logic.
We can view the set of statements in an argument as a kind of ``advice''
about which propositions or quantities to pay attention to.
If we understood how to propagate our uncertainty from one quantity
to another in a circuit
then we could plausibly apply similar ideas to propagate
uncertainty within a more complex argument.
Conversely if we are unable to produce coherent
probability estimates for circuits then it seems unlikely
that we can produce reasonable probability estimates
for the statements arising in a complex argument.

In particular,
trying to heuristically evaluate circuits forces
us to formalize and generalize the ``presumption of independence.''
For example,
we need to handle cases where we have information
about the \emph{pairwise} interactions between $x_1$, $x_2$, and $x_3$,
and want to make a guess about the conjunction $x_1 \wedge x_2 \wedge x_3$.
Generalizing the presumption of independence
in a coherent way appears to be quite challenging,
and we think it is the largest difficutly separating the
formalization of proofs from the formalization of heuristic arguments.

\section{Related work}\label{related}

There are many examples of probabilistic heuristic arguments
in the literature across a very wide range
of domains (e.g. \cite{cramer, cavity, diophantine, erdos, gameoflife}),
and many discussions of the philosophy of applying
heuristic arguments to unprovable statements (e.g. \cite{conway, dyson}).
But we are aware of very little work on formalizing these standards or attempting
to investigate heuristic arguments formally.

The most similar presentation we have encountered is the blog post \cite{tao},
which discusses the idea of assigning probabilities
to deterministic claims and presents two ``probabilistic heuristics:''
\begin{description}
    \item[Basic heuristic]
        ``If two or more of these heuristically probabilistic events have no obvious reason to be strongly correlated to each other,
        then we should expect them to behave as if they were (jointly) independent.''
    \item[Advanced heuristic]
        ``If two or more of these heuristically probabilistic events
        have \emph{some} obvious correlation between them,
        but no further correlations are suspected,
        then we should expect them to behave as if they were conditionally independent,
        relative to whatever data is causing the correlation.''
\end{description}
We are not proposing any revisions to these heuristics.
The main difference is that we are optimistic about capturing
them as part of a more general formal framework;
in this document we try to state that meta-problem.
In Appendix~\ref{cumulantappendix} we describe
our best attempt to design such a general framework
and explain why we consider it inadequate.

There are other types of reasoning that are distinct from heuristic arguments,
but close enough to be worth distinguishing specifically.

\textbf{Random models in number theory.} The closest thing to a formalization
of heuristic argument is the explicit use of random models as surrogates
for complex objects.
Most famous is the Cram\'er model of the primes \cite{cramer}, which suggests
that a statement is likely to be true of the primes
if it is true with high probability for a random set in which
each integer $x > 1$ is included with probability $\frac 1{\ln x}$.
Similarly, Erd\H{o}s and Ulam analyze Fermat's last theorem by proving
that the analogous statement would
almost surely be true if we replaced the perfect $n^{\text{th}}$ powers
with a random set of similar density \cite{erdos}.
We are unsatisfied by these arguments for a few closely related reasons:
\begin{itemize}
    \item Each such model applies to a relatively narrow range of questions---we
        are interested in finding more general rules that could be used
        to evaluate a wide range of questions
        (ideally across a wide range of domains).
        To do so, we would like to derive principles like the Cram\'er model
        from simpler principles,
        rather than including them in a very long list of ``heuristic axioms.''
    \item Even within a domain,
        the applicability of these random models
        is usually evaluated by informal judgment.
        For example, the Cram\'er model is usually
        considered to be applicable only for ``global''
        questions in an informal sense \cite{cramervcramer}.
        We would like to formalize this judgment of applicability,
        and capture it in a concrete heuristic estimator.
    \item Even when such models apply,
        we need to consider correction terms in order
        to get accurate estimates---for example the actual density of twin primes is
        about $30\%$ higher than the estimate from the Cram\'er model.
        How do we formalize the process for making
        this kind of correction without allowing
        the model to produce arbitrary conclusions?\footnote{
            For example, if we extend the Cram\'er model by allowing
            an argument to prove \emph{any} property $\phi$ of the primes
            and then treating the primes as a random set satisfying
            that property, then we can trivially produce
            arbitrary conclusions.
        }
    \item Beyond these difficulties, we are interested
        in formalizing the many heuristic
        arguments which are not captured by any such random surrogate
        (including the arguments about billiard balls and SHA-256
        discussed in Section~\ref{otherexamples}).
\end{itemize}

\textbf{Interactive proofs.} There is a large literature
exploring protocols by which powerful provers can convince bounded
verifiers of complex claims even in cases where there is no short 
traditional proof
(for the introduction of this concept see \cite{interactiveproofs}).
However none of these systems capture the kind of informal
heuristic arguments we discuss in this document,
and they often require extraordinarily powerful provers.
For example, while there are known interactive proof
systems that allow us to efficiently verify any statement
that has an exponentially-long proof,
these systems require the \emph{prover} to do exponential computation.
Heuristic estimators can be viewed as a type of interactive proof system
with very weak guarantees, but which are hopefully able to produce
reasonable estimates for realistically limited provers.

\textbf{Formalizations of logical uncertainty.}
Several authors have explored mechanisms for assigning probabilities
to arbitrary sentences of logic
(e.g. \cite{gaifman, hutter, demski, logicalinductors}).
However these approaches have primarily focused on establishing
coherence conditions
and on capturing \emph{inductive} reasoning,
i.e. ensuring that a reasoner eventually successfully predicts $\phi\of{n}$
given observations of $\phi\of{1}, \phi\of{2}, \ldots \phi\of{n-1}$.
These systems would not automatically recognize
intuitively valid heuristic arguments,
e.g. they would
not revise the probability they assign to the twin prime conjecture
after noticing the heuristic argument presented in Section~\ref{twinprimes},
although they would eventually \emph{learn} to trust these arguments
after observing them producing good predictions in practice.\footnote{Similarly,
    a neural network trained to predict the truth of mathematical
    statements may eventually learn to be a good heuristic estimator,
    but our goal is to understand \emph{what such a model learns} rather
    than to describe the process for learning.
    (Though as discussed in Appendix~\ref{alignment},
    and our primary interest is in using heuristic arguments to reason \emph{about}
    neural networks,
    rather than expecting them to capture the kind of reasoning performed \emph{by} neural networks.)
} Indeed, we can view ourselves as reasoners in exactly this situation,
trying to understand and formalize a type of reasoning
that appears to often make good predictions in practice.
Formalizations of inductive reasoning
may help clarify the standards we should use for evaluating a proposed
heuristic estimator,
but do not constitute a good heuristic estimator themselves.

\section{Conclusion}

Heuristic arguments based on the presumption
of independence often converge to empirically
reasonable estimates and can be intuitively compelling,
yet there is no existing formal framework for representing
or validating this kind of reasoning.
In this paper introduced a simple definition of a ``heuristic estimator,''
and stated a few open problems:
\begin{itemize}
    \item Finding any estimator that satisfies
        the desiderata in Section~\ref{desiderata}.
    \item Formalizing as many intuitively valid heuristic
        arguments as possible (Section~\ref{goal}).
    \item Finding better heuristic estimators for the output probability $\pc$ of a logical
        circuit $C$ (Section~\ref{circuitsection}).
\end{itemize}
Formalizing the previously-informal notion of proof played a central role
in modern mathematics and computer science,
and in the best case formalizing heuristic arguments
could open up analogous intellectual territory.
If successful,
it may also help improve our ability to verify reasoning about complex questions,
like those emerging in modern machine learning,
for which we expect formal proof to be impossible.

We have given a high-level overview of the questions
we find most exciting.
In the appendices we explore
heuristic arguments in more depth:
\begin{itemize}
    \item In Appendix~\ref{examples}
        we provide three additional examples of heuristic arguments
        to illustrate the breadth of applicability,
        highlight some important subtleties,
        and provide test cases for the open problem presented in Section~\ref{goal}.
    \item In Appendix~\ref{explaining} we introduce a distinction between
        ``inductive'' and ``deductive'' arguments,
        and explain why we believe probabilistic heuristic arguments may
        help capture the \emph{reason why} a statement is true.
    \item In Appendix~\ref{reasons} we present the strong conjecture
        that any true mathematical sentence has a deductive heuristic argument
        for its plausibility.
    \item In Appendix~\ref{cumulantappendix} we present a formalization
        of the presumption of independence
        in terms of the joint cumulants of several variables.
        We use this to define
        cumulant propagation,
        a simple heuristic estimator for the expected output of an arithmetic
        circuit with Gaussian inputs,
        and explain why we find this estimator inadequate.
    \item In Appendix~\ref{cherrypicking} we explore some examples
        where cherry-picking prevents heuristic estimators
        from converging to reasonable estimates in finite time.
    \item In Appendix~\ref{alignment} we briefly discuss some
        potential applications of heuristic arguments in machine learning.
\end{itemize}

\bibliographystyle{amsalpha}
\bibliography{references}

\appendix

\section{Examples of heuristic arguments}\label{examples}

\subsection[flt]{Fermat's last theorem\footnote{This heuristic
argument for Fermat's last theorem is standard,
essentially the same as the one appearing in \cite{erdos} and \cite{tao}.}}\label{flt}

\begin{question}
    For which integers $n$ does the equation $a^n + b^n = c^n$ have any solutions with $a, b, c > 0$?
\end{question}
We will start by asking: for a given $a > 0$,
how likely is it that there is a solution $a^n + b^n = c^n$ with $b \leq a$?
Equivalently
we can ask: is there an $x \in (a^n, 2 a^n]$
that satisfies both $\exists b \colon x = a^n + b^n$
and $\exists c \colon x = c^n$?

It is easy to calculate the probability that a random $x \in (a^n, 2 a^n]$ is of the form $a^n + b^n$:
there are $a^n$ numbers in the interval
and exactly $a$ numbers of the form $a^n + b^n$,
namely $a^n + 1, a^n + 2^n, \ldots, a^n + a^n$.
So the probability that a random $x \in (a^n, 2a^n]$
is of this form is $\frac {a}{a^n}$.

\newcommand{\afloor}{\left \lfloor \of{\sqrt[n]{2} - 1} a \right \rfloor}

Similarly,
there are $\afloor$
numbers of the form $c^n$,
namely $(a+1)^n, (a+2)^n, \ldots, \lfloor a \sqrt[n]{2}  \rfloor^n$.
So the probability that a random $x \in (a^n, 2a^n]$
is of this form is $\frac {\afloor}{a^n}$.

It is very hard to calculate the probability that both of these events
happen at once,
but we can apply the presumption of independence and estimate:
\begin{align*}
    \Pof{x}{\exists b \colon x = a^n + b^n \wedge \exists c \colon x = c^n}
    &\approx \Pof{x}{\exists b \colon x = a^n + b^n}\Pof{x}{\exists c \colon x = c^n} \\
    &= \frac {a}{a^n} \times \frac{\afloor}{a^n} \\
    &= \frac {\afloor}{a^{2n - 1}}
\end{align*}
Note that for $n > 1$ this probability is less than $\frac 1{a^n}$,
and so it cannot be the real probability for a randomly chosen $x$
(which will be $\frac 1{a^n}$ as long as there is even a single example).
We are unsure whether the true probability
is larger or smaller,
and this number reflects our uncertainty
both about the random choice of $x$
but also about how the $\nth$ powers are distributed in the interval.
This is an immediate consequence
of the presumption of independence.

Now we want to estimate the probability that there is \emph{any}
$x$ satisfying both properties.
For that purpose we apply the presumption of independence again,
and treat each event of the form
$\of{\exists b \colon x = a^n + b^n} \wedge \of{\exists c \colon x = c^n}$
as independent from the others
That gives us an estimate of:
\begin{align*}
    \P{\exists b, c \colon a^n + b^n = c^n}
    &= \P{\exists x \colon \of{\exists b \colon x = a^n + b^n} \wedge \of{\exists c \colon x = c^n}} \\
    &= 1 - \P{\forall x \colon \neg\of{\of{\exists b \colon x = a^n + b^n} \wedge \of{\exists c \colon x = c^n}}} \\
    &\approx 1 - \of{1 - \frac {\afloor}{a^{2n-1}}}^{a^n} \\
    &\approx \frac {\afloor}{a^{n-1}}
\end{align*}

Finally we want to calculate
the probability that this event occurs for \emph{any} $a > 0$.
To do this we apply the presumption of independence one last time:
\begin{align*}
    \P{\exists a, b, c \colon a^n + b^n = c^n}
    &= 1 - \P{\forall a \colon  \neg \exists b, c \colon a^n + b^n = c^n} \\
    &\approx 1 - \prod_{a = 2}^{\infty} \of{1 - \frac {\afloor}{a^{n-1}}}
\end{align*}
Approximating the infinite product we get:
\begin{center}
    \begin{tabular}{c|c}
        $n$ & $\P{\exists a, b, c \colon a^n + b^n = c^n}$ \\
        \hline
        \hline
        $2$ & $1$ \\
        \hline
        $3$ & $1$  \\
        \hline
        $4$ & $2.8\%$  \\
        \hline
        $5$ & $0.14\%$ \\
        \hline
        $\sum_{n \geq 6}$ & $0.005\%$
    \end{tabular}
\end{center}
So we expect there to be a solution for $n \leq 3$
and for there to probably be no solution for $n > 3$.\footnote{
    Note that the heuristic argument is wrong about the case $n = 3$,
    see Section~\ref{flt3case}.
}
This is because the expected number of solutions is roughly $\sum \frac 1{a^{n-2}}$
which diverges for $n=2$, diverges very slowly for $n = 3$,
and converges for $n > 3$.

These estimates are all defeasible and so are subject
to revision, even the estimates of $100\%$.
The numbers in this table reflect the uncertainty from \emph{chance},
but not the prospect of finding new considerations
that show that there is a correlation between the events $x = a^n + b^n$ and $x = c^n$,
or between the existence of solutions for different values of $a$.

\subsubsection{Checking small cases}

We estimated a $2.8\%$ chance that $\exists a, b, c \colon a^4 + b^4 = c^4$.
This was implicitly made up of intermediate estimates like
a $0.5\%$ chance that there is a solution with $b \leq a = 6$.
A very simple way that we could revise this probability is by checking some of those concrete
intermediate estimates,
e.g. by checking whether there is any $b \in \set{1, 2, 3, 4, 5, 6}$ such that $6^n + b^n$
is a perfect $4^{\text{th}}$ power.
If we find one then our probability of a solution will immediately go up to $100\%$,
and every time we fail to find one our probability of there
being any solution will go down slightly.

In fact there is only a single perfect $4^{\text{th}}$
power in the interval $(6^4, 2 \times 6^4]$, which is $7^4 = 2401$.
The difference $7^4 - 6^4$ is strictly between $4^4$ and $5^4$.
So $a^4 + b^4 = c^4$ has no solutions with $b \leq a = 6$.

This is technically a failure of our presumption of independence:
it turned out that the events $x = 6^4 + b^4$ and $x = c^4$
were anticorrelated.
This anticorrelation need not be for any deeper
reason---indeed we assigned this outcome a $99.5\%$ probability.
It could just be because there are only finitely many $x$
and it just so happened that none of them satisfied both properties.

After making this correction, our probability for any solution with $n = 4$ falls from $2.8\%$ down to $2.3\%$.
Checking small cases like this can quickly make us very confident that there are no solutions to $a^n + b^n = c^n$
for \emph{any} $n > 3$,
because the probability of a solution existing decays very rapidly as $a$ and $n$ grow.
After checking a modest number of cases we conclude that there is a $99.99\%$ chance that there are no solutions.
Of course this estimate is still defeasible,
just like our $100\%$ estimates for $n = 2$ or $n = 3$.

\subsubsection{Correlations between solutions for different values of $a$}

We assumed that the events $\exists b, c \colon a^n + b^n = c^n$
were independent for different values of $a$.
If those events were positively correlated then there could be a smaller
probability that one of them is true,
and if they were negatively correlated there could be a (slightly) larger probability.

In fact there is at least one obvious and important correlation:
for any $k > 1$,
we have
\[ a^n + b^n = c^n \Leftrightarrow (ka)^n + (kb)^n = (kc)^n.\]
Turning this around, if $a^n + b^n = c^n$
and $k > 1$ divides both $a$ and $b$,
then $\of{\frac ak}^n + \of{\frac bk}^n = \of{\frac ck}^n$
is also a solution.

So if we condition on not having found any solutions with $a' < a$,
that means that $x$ \emph{cannot} simultaneously be of the form $a^n + b^n$
and $c^n$ unless $x$ is relatively prime to $a$.
On average\footnote{We actually care about a particular weighted sum of values of $a$ that focuses
    on small values,
    and so we could make a more precise estimate here either by calculating exactly
    or by performing another heuristic estimate.
    But we can ignore these factors by presuming that $\phi\of{a}/a$ is independent of $1/a^{n-2}$.
}
this leaves us with only $\frac 6{\pi^2} a^n$ values of $x$ instead of $a^n$,
and so decreases our total estimate for the probability of a solution by $\frac 6{\pi^2} = 0.61\ldots$.

Now our estimate for $n = 4$ has fallen from $2.3\%$ to $1.4\%$.

\subsubsection{Correlations between $x = a^n + b^n$ and $x = c^n$}

We have assumed that the events $\exists b \colon x = a^n + b^n$ and $\exists c \colon x = c^n$
are independent if $a$ and $b$ are relatively prime.
But there are at least a few reasons for them to be correlated,
and as we notice these considerations we should change our prediction
for the probability that both of them occur:
\begin{itemize}
    \item Numbers of the form $a^n + b^n$ are not uniformly distributed over the range $(a^n, 2a^n]$,
        they are much more common closer to the bottom of the range.
        Similarly, numbers of the form $c^n$ are somewhat more common close to the bottom end of the range.
        A random number in the interval $(x - \epsilon, x+ \epsilon)$ has a probability
        of about $\frac 1{n x^{(n-1)/n}}$ of being a perfect $\nth$ powers
        (since the difference between consecutive $\nth$ powers in this range is roughly $n x^{(n-1)/n}$)
        and about $\frac 1{n (x - a)^{n/(n-1)}}$ of being of the form $a^n + b^n$.
        If we take the sum over a large number of intervals of this form,
        we converge to the estimate
        \[\int_{x = a^n}^{2 a^n} \frac 1{n x^{(n-1)/n}} \times \frac 1{n(x-a)^{(n-1)/n}} dx.\]
        Plugging $n = 4$ we get a number about $18\%$ higher than our previous estimate of $\frac{\sqrt[4]{2} - 1}{a^{n-2}}$,
        and so our estimate rises from $1.4\%$ to $1.7\%$.
    \item The $\nth$ powers are not not uniformly distributed modulo primes,
        and so this can introduce another correlation between the events $x = a^n + b^n$ and $x = c^n$.
        For example, every $4^{\text{th}}$ power is congruent to either $0$ or $1$ mod $5$.
        In order to get a more precise estimate for $\P{x = a^n + b^n \wedge x = c^n}$,
        we can compute:
        \begin{align*}
            &\P{\exists b \colon x = a^n + b^n =\wedge \exists c \colon x = c^n}\\
            &=\sum_{r} \P{x = r\; \mathrm{mod}\;{5}} \Pc{\exists b \colon x = a^n + b^n \wedge \exists c \colon x = c^n}{x = r\; \mathrm{mod}\;{5}} \\
            &\approx\sum_{r} \P{x = r\; \mathrm{mod}\;{5}}
 \Pc{\exists b \colon x = a^n + b^n}{x = r\; \mathrm{mod}\;{5}} \Pc{\exists c \colon x = c^n}{x = r\; \mathrm{mod}\;{5}}
        \end{align*}
        This sum leads to an estimate $4/3$ higher than our previous estimate\footnote{Keeping
            in mind that we also want to condition on no solutions with $a' < a$,
            and therefore only consider values of $x$ that are relatively prime to $a$.
        So we can separately consider the case where $a$ is divisible by $5$,
        in which $x$ is not divisible by $5$,
        and the case where $a$ is not divisible by $5$ such that $x$ is uniformly random mod $5$.}
        and so brings us from a $1.7\%$ probability of a solution up to $2.2\%$.
        There are similar adjustments for other divisors.
        which do not point in a consistent direction.
\end{itemize}

\subsubsection{The case $n = 3$}\label{flt3case}

We predicted that $a^n + b^n = c^n$ almost surely has a solution for $n = 2, 3$.
For $n = 2$ we predict a large number of solutions
and we can quickly find one, e.g. $3^2 + 4^2 = 5^2$.
For $n = 3$ we predict a very small number of solutions---we expect
about $1$ solution for $a, b \leq 100$,
$2$ solutions for $a, b \leq 10000$,
and $3$ solutions for $a, b \leq 1000000$.
But if we actually check all values up to a million,
we do not find any.
This is not decisive evidence
that we have made a mistake---we assigned this outcome a probability of about $1 / e^3 \approx 5\%$---but
it does suggest that something may be wrong.

We have already seen that there is one correlation
between the equation having solutions for different values of $a, a'$.
Taking that correlation into account only decreased the expected number of solutions by a factor of
$\frac 6{\pi^2}$,
but there are other more subtle correlations.

For example,
if $a^3 + b^3 = c^3$,
then we can compute that:
\begin{equation}\label{elliptic}
    \of{a^9 + 6 a^3 b^3 + 3 b^3 a^6 - b^9}^3 + \of{-a^9 + 3 a^6 b^3 + 6 a^3 b^6 + b^9}^3 = \of{3abc(a^6 + a^3 b^3 + b^3)}^3
\end{equation}
and hence a single solution generates an infinite family of solutions
by a second mechanism different from multiplying all of $a$, $b$, and $c$ by a constant $k > 1$.

This suggests that our independence assumption may break down.
In fact,
by doing some much more careful analysis
we can show that \emph{every} large solution to $a^3 + b^3 = c^3$
is generated by applying Equation~\ref{elliptic} to smaller solutions,
and hence if there are no solutions for small values of $a$
then there are no solutions at all.

This gives us a much more dramatic revision of a heuristic conclusion
than anything we had seen so far.
Observing Equation~\ref{elliptic} is much easier than proving Fermat's last theorem
(it was done centuries earlier)
but it is still extremely non-trivial
and causes us to revise the probability of a solution existing from $1$ down to $0$.

This revision is very distinctive to the equation $a^3 + b^3 = c^3$,
and typically when the naive heuristic suggests a very small number
of solutions this is correct.
For example, in 1769 Euler conjectured that there would also be no solutions
to 
\[a^4 + b^4 + c^4 = d^4.\]
In this case our basic heuristic argument again predicts
that there should be infinitely many solutions but that they should be very sparse.
In fact Euler's conjecture was disproven in 1988 \cite{elkies}.
The smallest counterexample (from \cite{counterexample}) is:
\[95800^4 + 217519^4 + 414560^4 = 422481^4.\]

%
%

\subsection{Hamiltonian cycles}\label{hamiltonian}

A weighted directed graph $G$ is a set of vertices $V$
and an edge-weighting function
$\sig{E}{V \times V}{\mathbb{R}}$.
(we indicate that an edge is absent by taking $E\of{u, v} = 0$).
A \emph{Hamiltonian cycle} in $G$
is a cycle that passes through each vertex in $G$
exactly once,
and the \emph{weight} of a cycle is the product of the edge weights.

For any given graph $G$, we can ask:
\begin{question}
    What is the total weight of all Hamiltonian cycles in $G$?
\end{question}
Even approximating the total weight of all Hamiltonian cycles is an extremely
difficult problem.\footnote{Determining whether there are any cycles
    with non-zero weight is NP-hard,
    and if the weights can be positive or negative
    then even determining the sign of the total weight of Hamiltonian cycles
    is \#P-hard.
}
In this section we discuss heuristic estimators for this quantity.

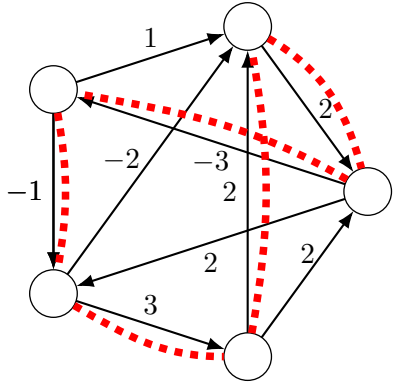
\begin{figure}
    \centering
        \begin{tikzpicture}
          \SetVertexNoLabel
          \begin{scope}
            \grEmptyCycle[prefix=a,RA=2/sin(60)]{5}
          \end{scope}
          \draw[thick,-{Latex[scale=1]}] (a4) -- (a0) node[midway, above] {$2$};
          \draw[thick,-{Latex[scale=1]}] (a3) -- (a4) node[midway, above] {$3$};
          \draw[thick,-{Latex[scale=1]}] (a3) -- (a1) node[midway, left] {$-2$};
          \draw[thick,-{Latex[scale=1]}] (a2) -- (a1) node[midway, above] {$1$};
          \draw[thick,-{Latex[scale=1]}] (a2) -- (a3) node[midway, left] {$-1$};
          \draw[thick,-{Latex[scale=1]}] (a2) -- (a3) node[midway, left] {$-1$};
          \draw[thick,-{Latex[scale=1]}] (a1) -- (a0) node[midway, right] {$2$};
          \draw[thick,-{Latex[scale=1]}] (a4) -- (a1) node[midway, left] {$2$};
          \draw[thick,-{Latex[scale=1]}] (a0) -- (a2) node[midway, below] {$-3$};
          \draw[thick,-{Latex[scale=1]}] (a0) -- (a3) node[midway, below] {$2$};
          \draw[red, dashed, line width=1mm] (a2) to[out=-80,in=80] (a3);
          \draw[red, dashed, line width=1mm] (a3) to[out=-30,in=180] (a4);
          \draw[red, dashed, line width=1mm] (a4) to[out=80,in=-80] (a1);
          \draw[red, dashed, line width=1mm] (a1) to[out=-30,in=105] (a0);
          \draw[red, dashed, line width=1mm] (a0) to[out=150,in=-10] (a2);
        \end{tikzpicture}
    \caption{A weighted directed graph is shown.
        If $u$ and $v$ are not connected by an arrow
        then $E\of{u, v} = 0$.
        A Hamiltonian cycle with weight $-1 \times 3 \times 2 \times 2 \times -3 = 36$
        is highlighted in red.
    }\label{circuit}
\end{figure}

\subsubsection{The naive estimate}

Let $W = \frac 1{n(n-1)} \sum_{u, v} E\of{u, v}$ be
the average weight of a randomly chosen edge from $G$
(including $0$ weights).
If we pick $n$ edges from $G$ independently at random,
then the expected product of their weights is exactly $W^n$.

There are $(n-1)!$ Hamiltonian cycles.
If we assume that the average weight of these cycles
is the same as the average weight of a random set of $n$
edges,
then the total weight of all Hamiltonian cycles is
\[ S_0 = (n-1)! W^n.\]
This corresponds to the presumption that if we pick $n$ edges
uniformly at random,
their total weight is uncorrelated with whether they are a Hamiltonian cycle.

\newcommand{\ecycle}{E_{\text{cycle}}}
\newcommand{\eunique}{E_{\text{unique}}}
\newcommand{\ein}{E_{\text{in}}}
\newcommand{\eout}{E_{\text{out}}}

\subsubsection{Estimates based on incoming or outgoing edges}

Every Hamiltonian cycle must have exactly one outgoing
edge from each vertex $u$.
So if we notice that a vertex $u$ has no outgoing edges
with non-zero weight
then every Hamiltonian cycle has weight zero,
and the probability of being a Hamiltonian cycle is \emph{not}
independent of the weight.
More generally,
if someone points out that certain vertices
have a very small weight of outgoing edges,
then it introduces a correlation between weight and being a Hamiltonian cycle.
We can incorporate this correction to get a more precise estimate.

Instead of picking $n$ edges at random,
we could pick an outgoing edge from each vertex at random.
Let $W_{u\rightarrow} = \frac 1{n-1} \sum_v E\of{u, v}$
be the expected weight of a random outgoing edge from $u$.
Then if we pick one outgoing
edge from each vertex,
the expected product of their weights is $\prod_u W_{u\rightarrow}$.

For $n$ edges chosen in this way,
we could again assume that the weight of the set is independent
of whether it is a Hamiltonian cycle.
If so, then the total weight of all Hamiltonian cycles would be:
\[ S_{\text{out}} = (n-1)! \prod_u W_{u \rightarrow}.\]
We can confirm empirically that this gives us a better estimator
for small random graphs. 

We could have done the same thing for incoming edges
instead of outgoing edges,
obtaining the estimate
\[ S_{\text{in}} = (n-1)! \prod_v W_{\rightarrow v}\]
where $W_{\rightarrow v} = \sum_u E\of{u, v}$.

\subsubsection[Combining the two estimators]{Combining $S_{\text{in}}$ and $S_{\text{out}}$}

The two estimates $S_{\text{out}}$ and $S_{\text{in}}$ can be very
different from each other,
and even have different signs.

This illustrates one of the core challenges
in constructing a heuristic estimator:
if we are given two different arguments $\pi_1$ and $\pi_2$,
how can we combine them to arrive at an estimate
that reflects all the information from both?
If there were cases without any intuitively
plausible way to do this kind of merging,
then it would provide a serious obstacle to our goal of defining
a heuristic estimator that aligns with our intuitions about validity.

In this case there turns out to be a relatively simple way to integrate
the estimates by applying the presumption of independence one more time.

First we will describe how to do this when all of the edge
weights are positive,
and then describe how to adapt it to handle negative weights.

We imagine selecting a sequence of $n$
(not necessarily distinct) edges each with probability
proportional to its weight.
Let $\ecycle$ be the event that these edges form a cycle.
The total weight of Hamiltonian paths is exactly equal
to $n^n(n-1)^n W^n \P{\ecycle}$,
and so we can restate our goal as estimating $\P{\ecycle}$.
Our original presumption of independence
was that $\P{\ecycle} \approx \frac{n^n (n-1)^n}{(n-1)!}$,
the same as if we had picked the edges uniformly at random.

We can also consider the events $\eout$ and $\ein$
that our random set of edges has exactly one outgoing
edge from each vertex or exactly one incoming edge to each vertex.
We know that $\ecycle \subset \eout \cap \ein$.

The estimate $S_{\text{out}}$ improves upon $S$
by exactly computing $\P{\eout}$
and then assuming that $\Pc{\ecycle}{\eout} = \frac {(n-1)!}{(n-1)^n}$
(which is equivalent to saying that for a uniformly random
set of edges satisfying $\eout$,
the product of the weights is independent from whether the edges form a cycle).
Similarly, $S_{\text{in}}$ computes $\P{\ein}$ and then assumes $\Pc{\ecycle}{\ein} = \frac{(n-1)!}{(n-1)^n}$.

We could get a better estimate if we could exactly compute $\P{\ein \cap \eout}$,
and then assume that $\Pc{\ecycle}{\ein \cap \eout} = \frac 1n$
(which is equivalent to saying that for a uniformly random set of edges satisfying
$\eout \cap \ein$,
the product of weights is independent from whether the edges form a cycle).

We cannot compute $\P{\eout \cap \ein}$
but once we are looking at the problem this way
it is easy to apply the presumption of independence
again to estimate $\P{\ein \cap \eout} \approx \P{\ein}\P{\eout}$.

Putting it all together, this gives us the estimate
\begin{equation}\label{sinout}
S_{\text{in+out}} = (n-1)! \frac{S_{\text{in}} S_{\text{out}}}{S_0}.
\end{equation}

This estimate is based on assuming independence
for edges sampled with probability proportional to their weight,
so it only applies if all edge weights are positive.
To generalize,
we can consider the estimates $S_{\text{in}}^+, S_{\text{out}}^+, S_0^+$ consisting
only of cycles where the product of edge weights is positive,
and compute $S_{\text{in+out}}^+$ using the analog of Equation~\ref{sinout}.
Separately we can compute $S_{\text{in+out}}^-$ consisting
of only terms where the product of edge weights is negative,
and then define $S_{\text{in+out}} = S_{\text{in+out}}^+ - S_{\text{in+out}}^-$.
It turns out to be straightforward to compute all of these quantities,
although this methodology can lead to particularly large multiplicative errors
in cases where $S_{\text{in+out}}^+ \approx S_{\text{in+out}}^-$
(as expected given that the problem is \#P-hard).

\subsubsection{The $\eunique$ correction}

If we evaluate the estimator $S_{\text{in+out}}$,
we find that it is better than either $S_{\text{in}}$
or $S_{\text{out}}$ for many distributions over graphs.
But there are some natural distributions (like power law distributed weights)
where it is actually much worse than even the naive estimator $S_0$.
One of our motivating beliefs,
discussed in more detail in Section~\ref{reasons},
is that \emph{whenever} we observe this kind of empirical failure
it means there is some heuristic argument that we are overlooking.

In this case the story is relatively simple.
We assumed that the events $\ein$ and $\eout$
were independent if we sample sets of edges with probability
proportional to their weights.
But these two events have an obvious correlation:
we are sampling sequences of edges with replacement,
and if we pick the same edge twice then \emph{neither} of these events will be true.
When the distribution of edge weights is heavy-tailed,
this is a very common reason for $\ein$ or $\eout$ to fail,
and so the independence assumption is badly wrong.

Let $\eunique$ be the event that no edge appears twice.
Rather than assuming that $\ein$ and $\eout$ are independent,
we would like to assume that they are conditionally independent
given $\eunique$.
That is, we would like to estimate:
\begin{align*}
    \P{\ein \cap \eout}
    &= \P{\eunique} \Pc{\ein \cap \eout}{\eunique} \\
    &\approx \P{\eunique} \Pc{\ein}{\eunique} \Pc{\eout}{\eunique} \\
    &= \frac{\P{\ein} \P{\eout}}{\P{\eunique}}.
\end{align*}
This suggests that we should multiply the estimator $S_{\text{in+out}}$
by $\P{\eunique}$.

Computing $\eunique$ exactly is not easy,
but we can again give a heuristic estimator for it.
For a given edge $(u, v)$,
the probability of $(u, v)$
appearing either $0$ or $1$ times in a random sequence
is exactly $(1 - p)^n + n p (1 - p)^{n-1}$,
where $p = \frac {E\of{u, v}}{W}$ is the probability that $(u, v)$
is picked at each step.

If we assume that these events are independent across all the edges $(u, v)$,
then we obtain an estimate for $\P{\eunique}$.
Multiplying $S_{\text{in+out}}$ by this estimate for $\P{\eunique}$
results in a new estimate for the sum of the Hamiltonian cycle weights,
and empirically we find that the resulting
estimate is typically significantly better than either $S_{\text{in}}$
or $S_{\text{out}}$.

Of course these events are not really independent (since if $(u, v)$
appears $0$ or $1$ times then it slightly decreases
the probability that another edge $(u', v')$ will appear $0$ or $1$ times),
but the assumption is quite close in many cases.
If a new heuristic argument led us to have a better estimate for $\P{\eunique}$,
then we would use that improved estimate instead.

\subsubsection{Considering concrete paths}\label{sumexample}

A very different way to heuristically argue
is to exhibit particular Hamiltonian cycles and compute their weight.

That is, we are interested in the sum
\[ S = \sum_{x \in \X} f\of{x} \]
where $\X$ is the space of Hamiltonian cycles and $f$ is the weight.

In the past sections we have shown a series
of increasingly sophisticated ways to derive a heuristic estimate
for the average value $f\of{x}$ over $\X$.
Write $\V{f, \pi}$ for our heuristic estimate of the average,
however we arrived at that estimate.

A particularly simple heuristic argument $\pi_x$
consists of a concrete $x$ together with a calculation of the value $f(x)$.
For a reasonable heuristic verifier,
we claim we should have:
\begin{equation*}
    \V{S, \pi, \pi_{x_1}, \ldots, \pi_{x_k}} =
    \abs{\X} + \sum_{i = 1}^k \of{f(x_i) - \V{f, \pi}}
\end{equation*}
That is, when $\tE$ sees that a particular value $f(x_i)$ is $\delta_i$
higher than it expected,
it increases its estimate for $S$ by $\delta_i$.\footnote{
    To be more precise,
    we really want to use $\tE$ to estimate the typical value of $f$
    on $\X \setminus \set{x_1, \ldots, x_k}$.
    In the case of Hamiltonian cycles this introduces an extremely small
    correction:
    if we have seen a single cycle,
    then each of the edges in that cycle only appears with probability
    $\frac 1{n-1} - O\of{\frac 1{(n-1)!}}$ amongst the remaining $(n-1)!$
    cycles.
    So instead of using a uniform distribution over edges
    we should revise all of our arguments to use this slightly non-uniform
    distribution.
    But this correction is very tiny unless $k$ is close to $(n-1)!$.
}

We think that $\tE$ should clearly change its estimate in this way.
You could also argue that it should change its estimate in a more fundamental way:
if $f(x_i)$ was higher than $\V{f, \pi}$,
it suggests that $\V{f, \pi}$ should be larger.
This is the underlying intuition behind a Monte Carlo estimate---the
random values we explore give us a reasonable indication of the
typical behavior of $f$,
and so we should update our estimate for $S$
based not only on the tiny number of values we saw explicitly
but on the assumption that unobserved values will behave similarly.

Roughly speaking,
we consider the linear contribution from $f(x_i) \rightarrow S$
to be a ``causal'' contribution,
roughly mirroring traditional deduction where we evaluate
individual factors that directly make a statement true or false.
In contrast,
we consider the contribution from $f(x_i)$ to $\V{f}$
to be an \emph{inductive} update,
where we change our beliefs about $\V{f}$
by observing evidence about $f$'s behavior and inferring
that there are likely to be common factors
that affect its behavior in every case.
We are particularly interested in \emph{deductive}
heuristic verifiers that do not make this kind of inductive update.
We explore this distinction in much more detail in Appendix~\ref{explaining}.

\subsection{Billiard balls}\label{ballexample}

\begin{question}
    Consider 15 perfectly elastic balls with radius 1 centimeter on
    a frictionless pool table measuring 1 meter by 2 meters.
    A line is drawn down the middle of the table dividing it into two 1 meter by 1 meter squares.
    Suppose that we choose the initial positions of the balls
    uniformly at random from the left half of the table,
    and we give each ball an initial velocity of 1 meter per second
    in a random direction.
    After 20 seconds,
    what is the probability that a majority of the 15 balls will be back on
    the left half of the table?
\end{question}

One way we could estimate the answer is by performing
a set of simulations with random initial conditions.
We find that unsatisfying for two reasons.
First, it performs badly if we want to get precise estimates
(e.g. for estimating probabilities very close to $0$
or very small biases away from a 50/50 chance).
More importantly but harder to formalize,
in Appendix~\ref{explaining} we try to explain
the intuitive sense in which a deterministic deductive
argument tells us something
that we do not learn from doing Monte Carlo estimate.

So in this section we will present
a deterministic but heuristic alternative to the Monte Carlo estimate.

\subsubsection{Stochastic differential equations}

The state of the table at any given time is described by $60$ numbers:
the $x$ and $y$ position and velocity of each of the $15$ balls.
It is easy to describe the initial configuration of the pool balls as a distribution
over this space,
but as time passes the probability distribution quickly becomes extremely
messy and has no short description.

One way we can track the evolution is by making a set of independence
assumptions in order to describe the this distribution more compactly.
The simplest independence assumption is
that all $60$ of these numbers are independent at any given time.
We will take a slightly more accurate assumption,
where we consider the correlation between a single ball's position
and velocity but assume that different balls are independent.

Under this assumption we need to track a distribution $p^t$ over
tuples $(x, y, \dot{x}, \dot{y}) \in \mathbb{R}^4$.
We'll define coordinates so that the table's walls are at $x = \pm 1$,
$y = 0$, and $y = 1$, with the left half of the table being the set $x < 0$.

Initially, $p^0$ has $(x, y)$ uniform over $[-1, 0] \times [0, 1]$
and $(\dot{x}, \dot{y})$ uniform over the unit circle.

If we ignore collisions between balls,
then the evolution of $p^t$ is very simple.
Over a short interval of time $\Delta t$, we have:
\begin{align*}
    x &\gets x + (\Delta t) \dot{x} \\
    y &\gets y + (\Delta t) \dot{y}
\end{align*}
If either of these positions goes outside of
the billiard table $[-1, 1] \times [0, 1]$,
we reflect it across the wall to put it back in the table
and we negate the associated velocity.

The collisions between balls introduce a much more
complex stochastic change to $\dot{x}$ and $\dot{y}$.
This is where we use the presumption of independence.
For a given pair of not-initially-overlapping tuples of positions
and velocities,
$B_0 = (x_0, y_0, \dot{x}_0, \dot{y_0})$ and
$B_1 = (x_1, y_1, \dot{x}_1, \dot{y_1})$,
it's easy to compute whether two balls with those parameters
would collide over the next $\Delta t$ seconds
and if so what the resulting velocities would be.
The limiting rate of collision as $\Delta t \rightarrow 0$ is 
\[ c\of{B_0, B_1} =
    \of{\dot{x}_0 - \dot{x}_1} \of{x_0 - x_1} + \of{\dot{y}_0 - \dot{y}_1}\of{y_0 - y_1}
\]
if $B_0$ and $B_1$ are exactly $2$ centimeters apart and $0$ otherwise.
If a collision occurs,
the new velocity for ball $0$ is $\of{\dot{x}_c(B_0, B_1), \dot{y}_c(B_0, B_1)}$,
where
\begin{align*}
    \dot{x}\of{B_0, B_1} &= \dot{x_0} + c\of{B_0, B_1} (x_0 - x_1) \\
    \dot{y}\of{B_0, B_1} &= \dot{y_0} + c\of{B_0, B_1} (y_0 - y_1)
\end{align*}

Given a probability distribution $p^t$ over $\mathbb{R}^4$
and a given tuple $B = (x, y, \dot{x}, \dot{y})$,
let $S$ be the set of tuples $B'$ that are just touching
$B$.
We can compute the limiting probability of a collision with another ball
in the next $\Delta t$ seconds, as $\Delta t \rightarrow 0$, as: 
\[ C^t(B) = 14 \int_{B' \in S}
p^t\of{B'} c\of{B, B'} dB'.\]
We've picked up a factor of $14$ because there are $14$
other balls with which any given ball could collide.

If a collision occurs,
the distribution over new velocities is
the distribution over $\of{\dot{x}\of{B, B'}, \dot{y}\of{B, B'}}$
for $B'$ sampled from $S$ with probability proportional to $p^t\of{B'}c\of{B, B'}$.
Again, this can be computed as a $3$-dimensional integral of $p^t$.

We now have a set of stochastic differential equations on $\mathbb{R}^4$
with jumps corresponding to collisions;
the presumption of independence
has reduced a $60$-dimensional problem
to a $4$-dimensional problem.
Although there are better approaches,
we can approximate the solution to such equations 
by the ``brute-force'' method of dividing $\mathbb{R}^4$
into $O\of{\frac 1{\epsilon^4}}$ small hypercubes with side length $\epsilon$
and tracking how each of them evolves over timesteps of length $\epsilon$.
This gives us an approximation to the final distribution
to within $O\of{\epsilon^2 t}$ error
in time $O\of{\frac {t}{\epsilon^6}}$.

Once we have a solution
in hand we can compute the probability $p$ that any given pool
ball is on the left half of the table.
We find that the result decays exponentially with $t$
and by 20 seconds it is extremely close to $\frac 12$.
Then to estimate the probability that most balls are on
the left half of the table we can apply
the presumption of independence again.

Note that this algorithm runs in time independent
of the number of pool balls,
and so we could have applied the same analysis to a set of $10^{20}$
gas molecules rather than $15$ pool balls.
For such systems even doing a Monte Carlo estimate would be intractable.

\subsubsection{Defeasibility}

These differential equations are only heuristically accurate,
and there could be important patterns that are
destroyed by the presumption of independence.

A simple example is that
if \emph{all} the pool balls are initially traveling almost exactly
straight up and down,
and if they start off with sufficiently
different $x$ positions, then they will stay on the left
half of the table and moreover they will never collide and
so never change their velocity.
It turns out that for large $t$ this possibility drives
most of the bias towards the left half of the table---for
large $t$ it suggests a bias of roughly $O\of{\frac {1}{t^{15}}}$,
whereas the bias from the estimate above
decays exponentially with $t$.\footnote{
    The bias drops off as $\frac 1{t^{15}}$ because
    there is a probability of $O\of{\frac 1t}$
    that any given pool ball is close enough to traveling perfectly
    up and down that it will remain roughly at the same $x$ coordinate
    for $t$ seconds.
    We need this event to occur for $15$ independent balls.
}
This possibility is completely ignored by the presumption of independence,
because it gives any given pair of balls a new independent
chance to collide in any given timestep.

As a much more exotic example,
this kind of heuristic estimate would give completely
wrong conclusions about a physical system that gives rise to complex life.
Thus \emph{proving} that any estimate of this form is accurate
effectively requires proving that the evolution of life is rare
in the system we're considering.
For interesting large systems that seems incredibly challenging,\footnote{
    It seems challenging to rule out even for a very large pool table.
    A small imbalance of pool balls towards the left half of the table
    provides a potential source of free energy,
    and while it seems difficult to build robust replicators
    out of pool balls
    we do not see how to rule out the possibility.
    (In this case it \emph{might} be possible to provably rule out life
    because the imbalance of pool balls is the only source of free energy
    and decays exponentially quickly,
    but if there had been any dynamics with longer timescales
    it no longer seems possible.)
    Note that the picture would be much simpler
    if we had initialized every pool ball randomly rather than
    restricting to the left half of the table.
}
and helps illustrate why proofs will typically be impossible.

\section{Inductive vs deductive arguments}\label{explaining}

\subsection{Proof vs evidence}\label{proofasexplanation}

Consider the problem of estimating $\pc$,
the probability that a circuit $C$ outputs $1$ for uniformly random inputs.
Rather than using a heuristic estimator,
we could use a Monte Carlo estimate:
draw some inputs $\set{z_i}_i$ at random and evaluate the empirical mean of $C\of{z_i}$.

If we test 1000 samples and find that $C\of{z} = 1$ for every one of them,
then that gives us strong evidence that $\pc$ is close to $1$.
In fact, this is much more convincing than a heuristic estimate
that $\pc \approx 1$, because there is no way we could have overlooked a key consideration.

Yet we find this estimate unsatisfying
and think it is still meaningful to look for a heuristic argument for $\pc$.
The Monte Carlo estimate gives us evidence that there is some structural feature of $C$
causing it to output $1$ most of the time,
but it doesn't help us see what that structure is---it doesn't show us \emph{why}
$C$ outputs $1$.
We are left with a mystery that we could explain by studying the circuit further.

In contrast,
we believe that a short \textbf{proof} that $\forall z \colon C\of{z} = 1$
would illuminate the relevant structure of $C$
and at least partially explain the phenomenon.
We don't know how to formalize this idea, but we can point to some
related observations:
\begin{itemize}
    \item A proof shows us what properties of $C$ led it to always output $1$,
        and so show us how we could change $C$ while preserving this property.
    \item We can make a Monte Carlo estimate regardless
        of how well we understand the circuit $C$,
        and in fact we could get
        the same estimate even if $C$ was cryptographically obfuscated.
        But efficiently proving that $C$ always outputs $1$
        requires de-obfuscating it.\footnote{
            Intuitively it shouldn't be possible to prove anything about an obfuscated circuit,
            but we can also prove this formally in the case of proving $\forall z \colon C\of{z} = 1$
            under indistinguishability obfuscation \cite{obfuscation} by using
            the ``punctured programming'' approach \cite{punctured}.
            Let $f$ be an indistinguishability obfuscator,
            such that it is computationally difficult to distinguish $f\of{C}$
            from $f\of{C'}$ whenever $C$ and $C'$ implement the same functionality.
            We'll show that we can't distinguish $f(C)$
            from a circuit that outputs $0$ on a single pseudorandomly chosen input,
            and therefore we can't prove $\forall z \colon f(C)(z) = 1$.

            Let $\sig{g}{\bits{m+1}}{\bits{m+1}}$
            be a one-way function,
            and define $C'(z) = 0$ if $g(0z) = 0^{m+1}$ and $C'(z) = C(z)$ otherwise.
            Then there is a half chance that $C' = C$,
            in which case we can't distinguish $f(C)$ from $f(C')$.
            But even if we are given $C'$,
            we can't tell the difference between cases where
            $g^{-1}\of{0^{m+1}}$ starts with $0$,
            in which case $C'$ is equal to $0$ on a single point,
            from cases where $g^{-1}\of{0^{m+1}}$ starts with $1$,
            in which case $C'$ equals $C$.
            So we also can't distinguish $f(C)$ from $f(C')$
            in cases where $C'$ outputs $0$ on a single input.
        }
        Obfuscation is an extreme case,
        but more generally it seems like proofs require
        us to identify the important structure in $C$
        rather than leaving it implicit.
    \item Intuitively proofs do often ``feel like'' explanations
        once we understand them,\footnote{
            There is a large philosophical literature on whether and when proofs are ``explanatory.''
            We don't intend to address the full complexity of that question,
            but just to make the much more mild claim that a proof is \emph{more}
            like an explanation than a Monte Carlo estimate is.
            A more precise statement of our view is that
            \textbf{short, constructive} proofs behave like explanations,
            but we won't defend even that weaker claim.
        } even if they are initially opaque.
        This is a relatively common intuition amongst mathematicians
        even if it lacks a clear philosophical basis,
        though note the important quantitative caveat about long proofs
        in Section~\ref{quantitativeexplanation}.
\end{itemize}

\subsection{Can heuristic arguments also be explanations?}

When discussing the difference between proofs and Monte Carlo
estimates it is tempting to focus on the certainty that proofs provide:
even if $C(z) = 1$ in 1000 random cases
the best we can say is that $\E{C}$ is probably not much less than $0.999$,
and it could even turn out that $\E{C} = 0.5$ and we just happened
to draw an extreme set of samples.

But the fact that proofs give us certainty
seems orthogonal to any of the advantages discussed in the last section.
The point is that a proof elucidates the structure of $C$,
not that it rules out the possibility of error.
If that's the case,
then a heuristic argument could potentially provide the same kind of elucidation
even though it doesn't provide the certainty.

Our intuition is that heuristic arguments based
on the presumption of independence
do show us ``why'' the corresponding statement is true.
For example,
we think that if the twin prime conjecture is true it is likely
to be ``because'' of the argument presented in Section~\ref{twinprimes},
and we should not necessarily expect to discover some further
facts about the distribution of primes.\footnote{One distinction
    with proofs is that we might find other structure about the primes
    that either makes the twin prime conjecture false
    or makes it true for a completely different reason.
    Sometimes this indicates that our initial explanation was wrong,
    but it can also be the case that a single claim has multiple
    sufficient explanations. For example, $A \vee B$ will often have two
    sufficient explanations, neither of which is wrong.
}

However, not all heuristic estimators have this property:
based on the definition of ``heuristic estimator'' in Section~\ref{verifiers},
a Monte Carlo estimator for $X$ would be an example
of a valid heuristic estimator.

So we would often like to restrict our attention
to a narrower class of heuristic estimators
that we will call \emph{deductive} estimators
which mirror the deductive structure of proofs (in contrast
with what we describe as the \emph{inductive} structure
of a Monte Carlo estimate).
Unsurprisingly we can't define this notion formally either.
But we can use it to inform the choice of examples
for the formalization problem posed in Section~\ref{goal},
and to guide our search for algorithms.

\subsection{Randomization does not capture this distinction}

%
Monte Carlo estimates are inherently random
while proofs are inherently deterministic.
So perhaps if we require a heuristic estimator
to be deterministic then we could
ensure that heuristic estimators have some of the same
explanatory benefit as proofs.

We think this doesn't work.
Consider a pseudorandom Monte Carlo algorithm
that estimates $\E{C}$ by evaluating $C$
at the values $f\of{0}, f\of{1}, \ldots, f\of{k}$
for some complicated and random-looking function $f$.

It is strongly believed that there exist formally
pseudorandom functions such that this pseudorandom Monte Carlo
estimate will also converge to the correct value
for \emph{every} circuit $C$.
Yet the pseudorandom Monte Carlo estimate tells
us no more than the random one did.
The failure to show \emph{why} $C$ outputs $1$
wasn't due to the use of randomness,
but due to the nature of the inference about $C$.

This leaves us searching for a better way to formalize
the difference between a Monte Carlo estimate and a proof.

%
\subsection{Induction vs deduction}

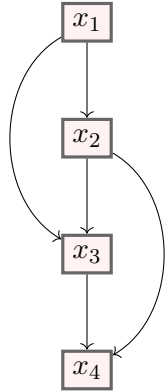
\begin{figure}
    \centering
    \begin{tikzpicture}[
        roundnode/.style={
            circle, draw=green!60, fill=green!5, very thick, minimum size=7mm
        },
        squarenode/.style={
            rectangle, draw=black!60, fill=red!5, very thick, minimum size=5mm
        },
        ]
        \node[squarenode] (x1) {$x_1$};
        \node[squarenode] (x2) [below=of x1] {$x_2$};
        \node[squarenode] (x3) [below=of x2] {$x_3$};
        \node[squarenode] (x4) [below=of x3] {$x_4$};
        \draw[->] (x1) -- (x2);
        \draw[->] (x1) to[out=210,in=150] (x3);
        \draw[->] (x2) -- (x3);
        \draw[->] (x2) to[out=-30,in=30] (x4);
        \draw[->] (x3) -- (x4);
    \end{tikzpicture}
    \caption{An illustration of a causal model. Arrows 
        represent a direct dependence of one variable on another. To fully specify
        a model, we would need to describe the domain
        of each variable and the conditional probability distributions
        $p\ofc{x_2}{x_1}$,
        $p\ofc{x_3}{x_1, x_2}$,
        $p\ofc{x_4}{x_2, x_3}$, and $p\of{x_1}$.}\label{causality}
\end{figure}

Our intuitive distinction between induction and deduction is
heavily informed by the analogy of reasoning in a causal model.
A causal model defines a probability distribution over a set of variables $\set{x_1, \ldots, x_T}$
by defining the conditional probability distributions for each variable $x_t$
given a set of values for the previous variables $\set{x_1, \ldots, x_{t-1}}$.
We often imagine the case where each variable  $x_t$
depends directly on only a few of the variables $x_i$ for $i < t$,
and is conditionally independent of the others;
we illustrate such a model in Figure~\ref{causality}.
Many reasoning problems
can be viewed as inferring the conditional probability distribution of a variable $x_i$
given the values of some other variables $\set{x_j}$.

We can divide up this inference problem into two parts:
\begin{description}
    \item[Forwards.] Given values for some early variables $\set{x_1, x_2, \ldots, x_t}$,
        we can repeatedly apply the conditional probability definition
        in order to compute the distribution of each of $\set{x_{t+1}, x_{t+2}, \ldots, x_T}$
        given $\set{x_1, \ldots, x_t}$.
    \item[Backwards.] If we're given a some later value $x_t$
        and want to infer the distribution over an earlier value $x_1$,
        then we need to also solve an inverse problem:
        for each way possible value of $x_1$ we compute $p\ofc{x_t}{x_1}$,
        and then we apply Bayes' rule to compute $p\ofc{x_1}{x_t} \propto \frac{p\ofc{x_t}{x_1}}{p\of{x_1}}$.
\end{description}
Most realistic problems require both kinds of reasoning.
For example, if I want to know $p\ofc{x_7}{x_4}$,
I need to infer the distribution over $\set{x_1, x_2, x_3}$ given the value of $x_4$,
and then use those to compute the distribution over $x_5$, then $x_6$, then $x_7$.

These two steps aren't always or even usually distinguished in inference algorithms,
but we still find it helpful to think of the two separately.
We think of the first as ``reasoning forwards'' from premises to conclusions,
in a way that closely mirrors the flow of logical implication in a proof.
We think of the second as ``reasoning backwards'' and trying to infer
the most likely explanation for some data.

In a causal model
we are working with \emph{bona fide} probability distributions,
whereas a heuristic estimator $\V{X}$
is working with its uncertainty about some deterministic quantity $X$.
Despite the differences,
we find the analogy to causal models helpful and we still expect
the same kind of ``forwards'' and ``backwards'' reasoning to occur
in realistic examples of reasoning about unknown but deterministic quantities $X$.

\begin{figure}
    \centering
    \begin{tikzpicture}[
        roundnode/.style={
            circle, draw=green!60, fill=green!5, very thick, minimum size=7mm
        },
        squarenode/.style={
            rectangle, draw=black!60, fill=red!5, very thick, minimum size=5mm
        },
        ]
        \node[squarenode] (def) {Definition of $C$};
        \node[squarenode] (C) [below=of def] {Facts about $C$};
        \node[squarenode] (x2) [below=of C] {$C\of{z_2}$};
        \node[squarenode] (x1) [below=of C, left=of x2] {$C\of{z_1}$};
        \node[squarenode] (x3) [below=of C, right=of x2] {$C\of{z_3}$};
        \draw[->] (def) -- (C);
        \draw[->] (C) -- (x1);
        \draw[->] (C) -- (x2);
        \draw[->] (C) -- (x3);
    \end{tikzpicture}
    \caption{Intuitively the structure of $C$ is ``logically upstream''
    of particular computations $C\of{z_i}$.}\label{logicalcausality}
\end{figure}
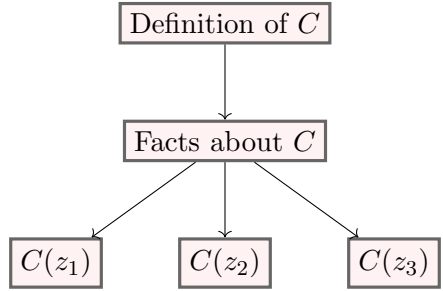

Now we can explain why we think a Monte Carlo
estimate for $\E{C}$ involves ``inductive'' reasoning.
The intuitive picture is illustrated in Figure~\ref{logicalcausality};
the circuit $C$ has a mathematical definition,
which logically entails some facts about $C$,
which in turn cause it to output $0$ or $1$ more often.
A Monte Carlo estimate doesn't try to discover
those underlying facts,
but instead observes various values $C\of{z_i}$
that are ``downstream'' of facts about $C$.
If it observes a bias then it implicitly infers that there
must have been some fact about $C$ leading to a bias,
and uses that to make predictions about new values $C\of{z_i}$.

We are instead interested in focusing on
what we will call \emph{deductive} heuristic estimators,
which deduce the relevant structural facts about $C$
directly from the definition,
rather than inferring their existence
from downstream consequences.

In the analogy to causal models,
a heuristic estimator is more like calculating the prior distribution
over $x_2$ by calculating $p\of{x_1}p\ofc{x_2}{x_1}$,
whereas a Monte Carlo estimate is more like observing $x_3$
and then doing a Bayesian update to adjust $p\ofc{x_2}{x_3}$
by the likelihood ratio $p\ofc{x_3}{x_2}$.

We expect realistic reasoning to involve
both this kind of ``forwards'' reasoning from premises
to conclusions,
and ``backwards'' Bayesian updating to adjust that prior based on observations.\footnote{Though
    in the case of Monte Carlo estimates,
    we can often obtain likelihood ratios so large that they completely
    overwhelm the prior.}
We are particularly interested in deductive heuristic estimators,
which try to isolate one part of this process,
for a few reasons:
\begin{itemize}
    \item We believe that less work has been put into
        formalizing the deductive part of the process,
        and the existence of simple arguments
        that are convincing but totally unformalized
        suggests that there may be significant low-hanging fruit for formalization.
        In contrast there is a much larger literature on approximate inference that
        focuses on the the inductive part of the problem.
    \item We think that it is very unlikely that there's a simple
        formalization of all reasoning about uncertain quantities.
        The purely deductive fragment seems much simpler and more likely
        to be governed by general principles (in analogy with logical deduction).
    \item We are interested in understanding ``why'' ML systems
        behave in a certain way,
        and tentatively hope
        that deductive heuristic estimators can shed some light
        on such questions for the reasons
        discussed at the beginning of this section.\footnote{
            We plan to discuss this hope in more detail
            in a forthcoming article,
            along with some recent examples of using this approach
            to solve problems in AI safety. We still have a lot of uncertainty
            but have some indication that this plan is coherent.}
\end{itemize}

\subsection{Example: estimating sums}\label{suminductive}

In Section~\ref{sumexample} we discussed heuristic estimators for a sum $S = \sum_{x \in \X} f(x)$,
and we considered heuristic arguments that simply compute $f\of{x_i}$ for concrete values $x_i$.

We think that such arguments should change the estimates for \textbf{both}
inductive and deductive reasons,
but the quantitative nature of the change is very different:
\begin{itemize}
    \item For a deductive heuristic estimator,
        learning that $f\of{x_i}$ is $1$ unit larger than we thought
        directly implies that $S$ will be $1$ unit larger than we thought---because
        $f(x_i)$ is one of the terms in $S$.
    \item If we are also reasoning inductively,
        then each value $f\of{x_i}$ also provides \emph{evidence}
        about the other values $f\of{x_j}$.
        Thus seeing 1000 random examples and finding they all have a value of $7$
        can lead to an estimate for $S$ of $7\abs{\X}$,
        even if $\abs{\X} = 2^{256}$ so that the direct impact
        of these examples is negligible.
\end{itemize}
The inductive reasoning generally leads to \emph{much} larger revisions.
But it also behaves qualitatively differently in several important ways:
\begin{itemize}
    \item The size of the inductive update depends a lot on how many
        examples we've already seen (and on our prior distribution over
        the behavior of $f$) whereas the size of the deductive
        update depends only on the difference between the observed value $f\of{x_i}$
        and our previous best guess $\V{f\of{x_i}}$.
    \item The inductive update depends on how the $x_i$ was chosen
        and whether it is representative of other values,
        whereas the deductive update depends only on the fact that $f\of{x_i}$
        appears in the sum $S$.
\end{itemize}

\subsection{Explanation seems to be quantitative}\label{quantitativeexplanation}

We can always prove $\forall z \colon C\of{z} = 1$
in a completely unenlightening way
by exhaustively checking every possible value of $z$.

In our view an exhaustive proof
is a valid deductive heuristic estimate,
and does constitute an explanation of the underlying phenomenon---we
just consider it a \emph{bad} explanation in a quantitative sense.
In this section we'll try to lay out some of the underlying intuitions
even though we can't formalize them.

An exhaustive proof has $2^n$ steps,
one for each input to $C$,
and each of these steps seems to work out ``by coincidence.''
We started out with a mystery of why $\forall z \colon C\of{z} = 1$
was true despite having heuristic probability $2^{-n}$;
but now we have the mystery of why every one of the $2^n$ steps
of the proof happened to work out.
The proof was no less surprising than the phenomenon-to-be-explained
and we've made no progress.

Given an explanation $\pi$ of a phenomenon $\phi$,
we can ask
\emph{how surprised we feel in total} after seeing the explanation---including
both how surprising the property now seems (measured by $\V{\phi, \pi}$)
as well as how surprised we are by the existence of $\pi$ itself.

We don't know how to quantify how surprising $\pi$ is,
but intuitively it is closely related to length: some steps of $\pi$
will involve coincidences,
and we effectively want to sum up surprisingness across those steps.
If we neglect the subtleties and just treat \emph{every} step as
surprising,
then we could define the quality of an explanation $\pi$ to be:
\[ \log \V{\phi, \pi} - \abs{\pi}.\]
This picture roughly mirrors an evaluation of a Bayesian
hypothesis as the log prior probability plus the log likelihood of the data.
This exact form seems unreasonable since $\abs{\pi}$
doesn't capture nuances in how surprising $\pi$ is,
but it seems like some more sophisticated
formula along these lines could give us a sense
of how well a heuristic argument $\pi$ explains the phenomenon $\phi$.

\section{Soundness}\label{reasons}

Suppose that we empirically discovered that
after some point the twin primes simply stopped appearing at the expected rate.
That is, we start checking the primes $p_1, p_2, \ldots$ greater than $N$,
and we find that $p+2$ is a composite in every single case we check.

After checking $10 \log{N}$ candidate primes and not finding any twins
we think that something is likely wrong;
we assigned a probability of only $0.005\%$ to seeing a stretch
this long without any twin primes.
After $100 \log{N}$ examples our probability is down to
$0.000000000000000000000000000000000000004\%$.


Of course we should not keep betting
that $p+2$ will be prime with probability $\frac 1{\log p}$.
At some point the inductive inference clearly trumps
the deductive heuristic argument
and we should not expect to see more twin primes.
But this raises the question:
was there \emph{necessarily} some argument
we overlooked,
some deductive heuristic argument
that would have revised our probability estimate
if we had noticed it?
Should we confidently expect that we'll learn something
if we keep investigating this phenomenon?

It seems implausible
that there could be no more twin primes ``by coincidence.''
But could it happen for a reason that is completely
beyond our understanding?

\subsection{Are all true statements heuristically plausible?}\label{soundnesssection}

\newcommand{\newsup}{\mathop{\smash{\mathrm{sup}}}}
\newcommand{\isv}{\inf \sup \V{\phi}}

For a given heuristic estimator $\tE$,
we say that a sentence $\phi$ is \emph{heuristically implausible}
if for any $\epsilon > 0$ there is a set of arguments $\pi_1, \ldots, \pi_n$
that can convince $\tE$ that $\Vpi{\phi} < \epsilon$,
and moreover such that there is no further set of arguments
$\pi_{n+1}, \ldots, \pi_m$ such that
$\V{\phi, \pi_1, \ldots, \pi_m} > \epsilon$.
Otherwise we say that $\phi$ is heuristically plausible.

That is, $\phi$ is heuristically plausible iff $\isv > 0$, where we define:
\[\inf \sup \V{\phi} = \inf_{\pi_1, \ldots, \pi_{n}}
\newsup_{\pi_{n+1}, \ldots, \pi_m} \V{\phi, \pi_1, \ldots, \pi_m}.\]

For example, the argument in Section~\ref{twinprimes}
implies that it is heuristically implausible
that there are only finitely many twin primes,
unless there is some further heuristic argument $\pi$
that the events ($x$ is prime) and ($x+2$ is prime)
are anticorrelated.

We'll say that a heuristic estimator is sound\footnote{
    We call this property \emph{soundness} in analogy
    with logical soundness because it means that if $\tE$
    is very confident about a statement, and nothing can change its mind,
    then the statement is true.
    By analogy we might use \emph{completeness} for the property that $\tE$
    eventually becomes confident about every true statement,
    which we discuss in Section~\ref{diagonalization}.
}
if every true statement $\phi$ is heuristically plausible.

This may look like a very weak statement
because we are only requiring $\tE$ to assign \emph{non-zero} probability
to true statements $\phi$.
Nevertheless,
asserting that a particular heuristic estimator is sound
can be an extremely strong statement.

For example,
suppose that $\phi$ is a computable property of natural numbers.
Unless the heuristic probability of $\phi\of{n}$ approaches $1$
sufficiently quickly for large values of $n$,
we heuristically expect $\P{\forall n \colon \phi\of{n}} = 0$.
So for any heuristically sound verifier
and any computable property $\phi$ that holds for all integers,
there must be a heuristic argument that $\P{\phi\of{n}}$ is extremely
close to $1$ for large values of $n$.

It's likely to be difficult or impossible
to prove that any interesting heuristic estimator is sound.
Proving this would require showing that there are never any ``grand coincidences''
that make a universally quantified statement true by chance alone.
But it's unclear what techniques could possibly
prove the absence of coincidences
for even a single sentence $\forall n \colon \phi\of{n}$.

Merely finding a deductive heuristic estimator which is \emph{plausibly} sound
would be extremely interesting.
We could summarize such a result as saying that ``everything happens for
a reason''---every true universally quantified statement
is explained by some heuristic argument accepted by $\tE$.

\subsection{Trivial forms of soundness}\label{uninteresting}

Some heuristic estimators $\tE$ are sound
for uninteresting reasons.

\textbf{Non-dogmatic.} If $\tE$ never assigns probability $0$
to any sentence unless it finds a disproof, then it will be sound
as long as the underlying proof system is sound.
Non-dogmatism seems like a reasonable epistemic principle,
and it is a defining property for many existing algorithms for assigning
probabilities to logical sentences
(e.g. \cite{gaifman, demski, logicalinductors, nonomniscience, hutter}).
We are interested in asking:
is non-dogmatism a \emph{fundamental} epistemic principle,
such that we should think of any sentence $\phi$
as having some probability of being true ``just because''?

\textbf{Easily persuadable.} Our definition
of plausibility requires that for every argument $\pi^-$
that $\phi$ has probability $0$,
there is a counterargument $\pi^+$ showing that $\phi$ has non-zero
probability after all.
This property is trivially satisfied if the ``second arguer wins,''
e.g. if $\tE$ simply defers to the longest argument it sees.
Soundness does not guarantee that an estimator is reasonable
or expressive in any interesting sense;
soundness is only interesting
for heuristic estimators that have other desirable properties.

\textbf{Moved by inductive evidence.} There is nothing in the definition
of a heuristic estimator
that prevents it from accepting arguments like:
``$\phi\of{n}$ has been
true for the first $10^{100}$ values of $n$,
so I'll give a 50\% chance to $\forall n \colon \phi\of{n}$.''\footnote{
    Though note that it is not easy to accept such arguments
while satisfying the desiderata in Section~\ref{desiderata}.}
If in fact $\forall n \colon \phi\of{n}$,
then it is possible
to exhibit an arbitrarily long list of positive examples.
Thus any heuristic estimator that accepts inductive
evidence is likely to be sound.
We only consider
soundness interesting for what we called \emph{deductive}
heuristic estimators in Section~\ref{explaining}.

\textbf{Moved by hypothetical reasons.} Even if we haven't yet found
any structure in the primes that could cause the twin prime conjecture to fail,
we think that a reasonable heuristic estimator could be open to heuristic arguments
about the probability that there exists some structure we haven't yet noticed.
This \emph{might} ensure that almost any statement is heuristically plausible,
if the estimator always holds out enough hope for the possibility that there
is a key structural fact that it hasn't yet noticed.
We are interested in asking:
was that hope justified---e.g.
was it \emph{actually} the case that if the twin conjecture fails
it's because there is a concrete reason for an anticorrelation?
Or could our heuristic estimator avoid assigning probability 100\%
only by forever holding out the possibility of seeing a hypothetical counterargument
$\pi$ that doesn't actually exist?

\subsection{Quantitative soundness}\label{quantitativecompletness}

So far we've considered the weak condition $\isv > 0$.
We expect an ideal heuristic estimator to assign true sentences
probability significantly more than zero---but exactly how much more?

On the one hand, we expect there to be true sentences of length $k$
that are assigned probability $O\of{2^{-k}}$.
For example,
let $X$ be the definition of an algorithmically random real number\footnote{
    For example Chaitin's Omega, the probability that a randomly
    chosen Turing machine halts.
    The key feature of such numbers is that for any computable
    process, the probability of guessing the first $k$
    bits of the number correctly provably decays like
    $c 2^{-k}$ for some constant $c$.
}
and consider the statement that the first $k$ bits of $X$
are $0.x_0x_1\ldots x_k$ for some particular $x_i \in \bits{}$.
One of these $2^k$ statements will turn out to be true,
but we don't expect any heuristic estimator to be
able to guess which one with probability better than chance.

On the other hand,
consider the set of true sentences $\phi$ with length
$\abs{\phi} < k$ such that\footnote{
    We'll consider the length $\abs{\phi}$ of
    a sentences when it is written in binary
    in some particular prefix-free encoding,
    i.e. a representation
    such that no syntactically valid sentence is a prefix of any other
    and hence $\sum_{\phi} 2^{-\abs{\phi}} \leq 1$.
} 
$\isv < 2^{-k}$.
There are at most $2^k$ such sentences.
And if $\tE$ is ``well-calibrated,'' then each of these sentences
ought to be true with probability less than $2^{-k}$.
Therefore in expectation at most $1$ of these sentences will turn out to be true.

In fact, the same argument suggests that there are expected
to be at most $\epsilon$ true sentences
of \emph{any} length
such that $\isv < \epsilon 2^{-\abs{\phi}}$.

This gives us a quantitative
version of the soundness condition from the last section:
\begin{equation}\label{soundness}\tag{$\epsilon$-soundness}
    \text{For all true sentences $\phi$:\,}
    \isv > \epsilon 2^{-\abs{\phi}}
\end{equation}
We expect a good verifier to satisfy this condition for a sufficiently
small constant $\epsilon$.
In fact this requirement is fairly likely to be true
even for $\epsilon = 1$,
but it is heuristically almost surely true as $\epsilon \rightarrow 0$.

\subsection{Empirical predictions}\label{empirics}

Mathematicians often discover initially-unexplained empirical regularities.
For example,
in 1853 Chebyshev \cite{chebyshev} observed that if you divide
a random prime number by $4$
the remainder is $3$ slightly more often than it is $1$---even though
we might have heuristically expected those two events to be equally common.
In fact it seems to be the case that
for almost all\footnote{
    The set of $N$ satisfying this property has logarithmic density
    more than 99\%.
    It is straightforward but slightly subtle to translate
    this into a true statement of the form $\forall n \colon \phi\of{n}$.
} $N$,
a majority of primes less than $N$ have remainder $3$ mod $4$.
After discovering this fact,
how confident should Chebyshev have been that mathematicians
would eventually find a clear explanation?\footnote{
    \cite{races} contains a clear discussion of this and other similar phenomena.
    In fact there is a heuristic argument that the prime powers ought to
    be uniform mod $4$,
    and hence that the primes themselves must be biased towards $1$,
    though we don't believe this argument was recognized for many decades
    after Chebyshev's observation.
    This can be derived from a generalization
    of the Riemann hypothesis,
    which also has no proof but is heuristically supported,
    see \cite{chebyshev-explanation}.
}

The existence of a sound heuristic estimator is closely related
to a more general empirical prediction about the practice of mathematics:
every time we find an empirical regularity like Chebyshev's bias,
we will eventually be able to find a concrete heuristic argument
explaining why the regularity occurs.

We don't have a concrete heuristic estimator so we can't evaluate
the claim formally,
but we can still ask whether mathematicians find an \textbf{informal}
heuristic argument.
Similarly,
we don't have infinite time so we can't ask whether
we will eventually find such an argument,
but we can ask whether we typically find them \textbf{quickly}.
For example we can ask:
how many observed empirical regularities
are currently unexplained despite a significant effort?
How reliably and quickly we can find an explanation
for a currently-unexplained empirical regularity
if we decide to investigate it thoroughly?

If we ask the analogous question for proofs the situation looks bleak:
most domains of mathematics are full of unproven conjectures
that are strongly believed.
Moreover it is not hard to spend an afternoon experimenting with numbers
to arrive at a novel conjecture that is probably true
but unlikely to be resolved even given decades of effort.

But if we consider heuristic arguments as well as proofs then it appears to us
that a significant majority of well-studied empirical
regularities have been adequately explained.\footnote{
    We don't believe that this observation is the result of a selection effect.
    Many researchers would consider
    a completely-inexplicable empirical regularity to be extremely interesting,
    and so we would expect potential counterexamples to this empirical
    regularity to be particularly \textbf{unlikely} to be forgotten.
}
Similarly, it seems quite challenging
and noteworthy to discover empirical regularities
that don't have a simple heuristic explanation.
And if a currently-unexplained regularity was selected and investigated thoroughly,
we believe it is very likely that an explanation could be found
within months or potentially years rather than decades.

In our experience it isn't controversial to suggest
that there almost certainly \emph{exists} an explanation for any given
empirical regularity.
The alternative, that such regularities can be fundamentally
inexplicable coincidences,
seems to be considered unlikely.
What is striking about the current situation is that
despite this historical pattern
we don't have any candidate formalization of what we mean by ``explanation.''
If this is really a robust pattern,
then that strikes us as a deep fact about the nature of mathematics,
and we expect that there is \emph{some}
better definition of explanation than ``a paper that leaves
mathematicians feeling convinced that the phenomenon is plausible.''

The best candidate counterexample we are aware of
is the consistency of strong axiom systems,
which we will discuss in Section~\ref{consistency-counterexample}.
Reasoning about ``explanations'' for consistency claims
is very subtle and probably requires having a more precise
definition of what constitutes  an explanation,
so for now we think it's hard to tell
whether consistency statements have plausibility arguments.
The empirical prediction discussed in this section
seems interesting even
if we explicitly set aside these cases.
As we discuss in Section~\ref{beyondgodel} we don't believe
that consistency statements
are the most important way in which proof systems are incomplete.

\subsection{Incompleteness and diagonalization}\label{diagonalization}

Soundness is the requirement that $\isv > 0$ whenever $\phi$ is true.
We could also consider completeness,
the property that $\isv = 1$ whenever $\phi$ is true
(or the even stronger property $\sup \inf \V{\phi, \pi} = 1$).

We don't think that we should expect such a strong principle to hold
even if $\tE$ is an ideal formalization of heuristic reasoning.
No matter how good we are at reasoning,
there are many complicated questions where we shouldn't expect
to get to a confident answer no matter how many arguments we see.

Unsurprisingly,
we can also show that this property is impossible via a diagonalization
argument. Define $G$ by quining such that:
\[ G \Leftrightarrow \inf \sup \V{G} < 1.\]
If $\inf \sup \V{G} = 1$,
then $G$ is false and hence $\neg G$ is a true statement
with $\inf \sup \V{\neg G} = 0$.
Thus it can't possibly be the case that $\isv = 1$ for every true
sentence $\phi$.

We are aware of no similar diagonalization obstruction to satisfying soundness.
Here are some examples of self-referential sentences $G$
and possible ways that a heuristic estimator could handle them
while being consistent with soundness:
\begin{description}
    \item[$G \coloneqq \sup \inf \V{G} = 0$.] We expect $G$
        to be true, and to have $\inf \sup \V{G} = 1$.
        No matter what argument $\pi$ you make suggesting that $G$ is true,
        there is another argument $\pi'$ suggesting that actually $G$
        is false, perhaps by pointing out that $\V{G, \pi}$ is large.
        We expect this process to go on forever
        and for $\tE$ to oscillate indefinitely.
        This is closely related to the examples in Section~\ref{cherrypicking},
        which give a simpler argument that we could not achieve
        the stronger form of soundness $\phi \Rightarrow \sup \inf \V{\phi} > 0$.
    \item[$G \coloneqq \inf \sup \V{G} = 0$.] We expect $G$
        to be false, and to have $\inf \sup \V{G} = 1$.
        This is a similar case where arguments cause $\tE$
        to oscillate indefinitely.
        In these cases, the ``innermost quantifier wins.''
    \item[$G \coloneqq \inf \sup \V{G} < 1$.] We expect $G$
        to be true, with $\inf \sup \V{G} = 1 - \epsilon$
        where $\epsilon > 0$ is $\tE$'s limiting probability that $\tE$ is unsound.
        Soundness is compatible with a model being uncertain about its
        own soundness.
    \item[$G \coloneqq \sup_{\pi} \V{G, \pi} < 1$.] We expect $G$ to be false
        with $\inf \sup \V{G} = 0$.
        This is an existentially
        quantified statement, so we expect there to be a simple
        argument $\pi^*$ such that $\V{G, \pi^*} = 1$.
        Computing $\pi^*$ is a simple proof that $G$
        is false, and hence $\inf \sup \V{G} = 0$.
\end{description}
It's not clear that a heuristic estimator should behave in these ways,
but these behaviors are consistent with soundness and we consider them
reasonable options suggesting that our goals for heuristic estimators
are mild enough to be compatible with self-reference.
Allowing models to be unsure about their own soundness,
and allowing their probabilities to oscillate indefinitely,
seem to avoid most plausible paradoxes.

\subsection{Consistency statements}\label{consistency-counterexample}

A central example of a true statement
that is unprovable in a theory $T$ is the consistency $\mathrm{Con}\of{T}$ 
of $T$ itself.
It's natural to wonder if it's possible to make a heuristic argument
that $T$ is consistent without needing to use axioms beyond $T$.
If it's not,
then this might be a counterexample to the empirical prediction
in Section~\ref{empirics}
and a way to show that interesting forms of heuristic soundness are unachievable.

Overall we'll argue that it's premature to try to answer this question---there
is no obvious diagonalization obstruction,
and there are plausible arguments for consistency,
but we can't really evaluate those arguments without
having a much clearer picture of how a hypothetical
deductive estimator would work.

This section ventures into even more ungrounded speculation,
and so we recommend that most readers skip it
unless they find the question particularly interesting.

\subsubsection{The problem}

\newcommand{\con}{\mathrm{Con}\of{\mathrm{ZFC}}}
To illustrate the issue,
let's work within ZFC\footnote{
    We are working with ZFC rather than a theory of arithmetic because
    it seems that set theory really is necessary in order to carry
    out the kind of intuitive argument that mathematicians make
    for the consistency of weaker systems---we believe
    that axiom systems are consistent because they have models,
    and so we need a theory rich enough to be able to talk about such models.
    Unfortunately using an expressive enough set theory to capture
    such arguments makes it even harder to think about how a hypothetical
    heuristic estimator might work.
}
and consider the statement:
\[ \con \coloneqq \forall \pi \colon \text{$\pi$ is not a proof of a contradiction in ZFC.} \]
There is a simple heuristic argument that $\con$ should be false:
there infinitely many possible proofs, and if we treat each of them
as having some independent chance of deriving a contradiction
then almost surely one of them will.\footnote{
    This comes down to a counting argument and isn't entirely clear.
    In particular, we need to consider the number of valid $n$-step proofs,
    together with the heuristic probability that a particular
    $n$-step proof derives a contradiction
    given that no smaller proof has.
    We won't discuss this argument,
    but we think that a reasonable heuristic estimator
    would probably conclude that any given set of axioms
    is almost surely inconsistent as a default
    until it sees something about the structure of the axioms
    that explains why they could be consistent.
}
So in order to be sound,
we need to find a heuristic argument $\pi$
that explains why $\con$ is plausible after all.

We'll consider two plausible ways that we could heuristically
argue for $\con$.

\subsubsection{Set-theoretic approach}

One way to argue that ZFC is consistent
is to find a transitive model for ZFC,
i.e. a set $M$ such that each axiom of ZFC is true
when the quantifiers range over $M$.
If we have such a model,
then we can inductively show that ZFC only proves statements $\phi$
that are true of $M$,
and hence that ZFC can't derive a contradiction.

At face value arguing for the existence of such an $M$
isn't necessarily any easier than arguing for $\con$:
the axioms of ZFC themselves specify an infinite list of claims
about $M$,
and so the existence of a set $M$ satisfying all of them
is heuristically implausible.

However, the axioms of ZFC have a special structure
that makes it plausible that we can satisfy them all.
In particular,
$M$ is a transitive model of ZFC if and only if it 
contains the integers and is closed
under a few fundamental operations---taking
finite collections, unions, power sets,
or forming a set $\setc{f(x)}{x \in S}$ for a function $f$ and a set $S \in M$.

ZFC is able to build very large sets that nearly
satisfy these properties by starting from the integers
and then iteratively adding more and more sets.\footnote{
    Technically we start with $V_0 = \emptyset$,
    define $V_{n+1} = \mathcal{P}\of{V_n}$,
    and define $V_\alpha = \cup_{\beta < \alpha} V_{\beta}$
    for each limit ordinal $\alpha$.
    ZFC is able to construct $V_{\alpha}$ for every ordinal $\alpha$,
    and can prove that $V_{\kappa}$ is a transitive model of ZFC
    whenever $\kappa$ is an inaccessible cardinal.
}
Using this idea it can prove that such a transitive model $M$
exists as long as there is an
\emph{inaccessible cardinal} $\kappa$:
a set bigger than any union\footnote{I.e. $\kappa$ is bigger than $\cup_{x \in S} x$
for any set $S$ smaller than $\kappa$ each of whose elements is smaller than $\kappa$.}
or power set of smaller sets.
We have no idea whatsoever whether the existence of an inaccessible
cardinal is heuristically plausible.
There are potential arguments on both sides,
but arbitrating the question seems impossible or meaningless given our current
state of uncertainty about how a hypothetical heuristic estimator would work.

%

The main point we want to make
is that the special structure of ZFC does appear to give us a concrete
reason to think that ZFC may be consistent
via the construction of the cumulative hierarchy.
This argument can be appreciated within ZFC
even if ZFC cannot establish the existence of an inaccessible
cardinal and therefore cannot tell whether the process goes on long
enough to actually produce a transitive model.
This leaves the heuristic status of the consistency of ZFC highly unclear,
even though we heuristically expect that \emph{most} sets of axioms are inconsistent.

\subsubsection{Explicit reflection principle}\label{selftrust}

We can formalize the idea of one proof system $T_1$
trusting another $T_2$, about a statement $\phi$, by asking whether
$T_1$ proves a theorem like: ``If $T_2$ proves $\phi$,
then $\phi$ is true.''
L\"{o}b's theorem \cite{lob} states
that if a proof system trusts itself about a statement $\phi$,
then it can immediately prove $\phi$.
Thus it's impossible for a proof system to trust itself
except when it already knows the answer.

What would it mean for a deductive heuristic estimator to trust itself?
Imagine that we find an argument $\pi$ which tells us not that $X$ is large
but that there \emph{exists} an argument $\pi'$ that $X$ is large.
If $\tE$ trusts itself,
then $\pi$ should already be enough to change its beliefs
about $X$---we shouldn't have to actually find the argument $\pi'$
and present it to $\tE$ explicitly.
We could imagine adding an explicit deduction
rule that allows $\tE$ to make this inference.

We can't really meaningfully investigate this kind of deduction rule
without having a much clearer picture of how a deductive
heuristic estimator might work.
But it's worth noting that L\"{o}b's theorem
and similar obstructions don't seem to apply,
and it seems plausible for a deductive heuristic estimator to trust itself
in this sense.
The key difference is that $\tE$ considers the existence of $\pi'$
for which $\V{X, \pi'}$ is large to be \emph{prima facie} reason
to believe that $X$ is large,
but this does \textbf{not} correspond to $\tE$ assigning high probability
to any material implication of the form
$(\text{$\V{X, \pi'}$ is large}) \Rightarrow (\text{$X$ is large})$.

If a heuristic estimator both accepted
the axioms of ZFC and trusted itself in this way
then it may be able to directly deduce
that the axioms of ZFC
are almost surely consistent:
\begin{itemize}
    \item For any finite set of axioms from ZFC,
        ZFC can prove that those axioms are consistent.
        Moreover, ZFC can \emph{prove} that for any finite set of axioms,
        there is a proof in ZFC that those axioms are consistent.
    \item If our heuristic estimator considers the mere existence
        of an argument to be persuasive,
        then proving that there exists an argument for every set of axioms
        is enough to infer that every set of axioms is almost surely consistent.
    \item There are only countably many sets of axioms, and so if our estimator
        knows that every one of them is almost surely consistent
        then it can conclude that it is almost surely the case that every one of them is consistent,
        and hence that ZFC itself is consistent.
\end{itemize}
This estimate is defeasible,
and e.g. if ZFC later found a proof of a contradiction then it would
of course conclude that ZFC wasn't consistent after all (though at that
point it would have bigger problems since it would be possible
to make convincing arguments for arbitrary claims).


\subsection{Other failures of proofs}\label{beyondgodel}

Although G\"{o}delian statements are the most famous failure of proofs,
it seems likely to us that unprovability is ubiquitous.
Our position on these questions is similar to the one expressed by Conway in \cite{conway}.

We think of G\"{o}delian statements as an interesting challenge case
for soundness of a heuristic estimator,
but we \emph{don't} think of proving G\"{o}delian statements as the central
way in which heuristic estimators overcome the incompleteness of proof systems.

For a more prosaic example of incompleteness,
take $\sig{f}{\mathbb{N}}{\bits{}}$ to be a complex function with no apparent
structure or bias towards $0$ or $1$.
Then consider the statement:
\[ \phi\of{f} \coloneqq \forall N > 100: \sum_{n = 1}^{N} f(n) > 0.01 N.\]
Heuristically this statement is almost surely true
and can fail only if $f$ has some special structure that we've overlooked.
But on the other hand,
it seems that $\phi\of{f}$ can only be proven if $f$ has special structure that can
be leveraged by a proof.
So a structureless $f$ would make $\phi\of{f}$ both true and unprovable.

How we can we reconcile this pessimistic view with the empirical
success of mathematicians at proving theorems?
\begin{itemize}
    \item If we write down a simple function,
        it is quite likely to have plenty of structure
        (even if there is no obvious structure at a first glance).
        Indeed, cryptographers spend a great deal of effort trying
        to find simple functions without any special structure
        that would make them amenable to cryptanalysis,
        and na\"{\i}vely choosing ``random-looking'' functions rarely succeeds.
        Writing down a concrete simple function $f$ for which $\phi\of{f}$
        is unprovable strikes us as a very similar challenge.
        That said, we expect many such functions to exist
        and to be extremely ``mundane,''
        looking more like cryptographic hash functions
        than self-referential sentences.
    \item Mathematicians systematically avoid areas
        without the kind of structure that facilitates proofs.
        For example the Collatz conjecture concerns a very simple function
        (much simpler than almost any function that has been found sufficiently ``structureless''
        to be usable in cryptography),
        and we could imagine a rich sister field to number theory proving simple statements
        about similar dynamical systems.
        But it doesn't exist in part because
        mathematicians have gotten very little traction on proving statements of this type.
        Number theory has flourished precisely because mathematicians
        have been able to say interesting things about the primes for thousands of years.
\end{itemize}
We believe these two facts largely explain the empirical success of proofs,
and are consistent with a perspective where unprovability is the ``default''
situation except when special structure makes proof possible.

Regardless of whether this perspective on unprovability is correct,
one special feature of G\"{o}delian statements
is that it is easy to \emph{prove} that they are unprovable (in a stronger theory).
In contrast, in the case of a typical ``structureless'' function $f$,
we expect it to be unprovable that $\phi\of{f}$ is unprovable.
So even if this kind of unprovability were ubiquitous,
G\"{o}delian statements would likely remain the prototypical
examples of unprovable sentences.
This mirrors the situation in complexity theory,
where it is suspected that ``generic'' functions cannot
be efficiently computed,
but diagonalization arguments
are practically the only source of \emph{provably} hard-to-compute functions.

\subsection{Quantitative bounds on argument length?}\label{argumentsoundness}

We are often interested in statements about strictly finite objects,
for example the claim that $\forall x \colon C\of{x} = 1$
for a particular circuit $\sig{C}{\bits{m}}{\bits{}}$.
In these cases heuristic soundness is trivial,
because there is a finite proof of the statement
by exhaustively considering every possible input $x$.

Nevertheless we would consider a heuristic estimator unreasonable
if $\forall x \colon C\of{x} = 1$
but the only way to heuristically argue for this fact
was to exhaustively consider every input.

Intuitively this is damning because
the fact that an exhaustive proof derives the conclusion
$\forall x \colon C\of{x}=1$
is \emph{itself} surprising---in fact just as surprising
as the original claim.

We discuss this idea informally in Section~\ref{quantitativeexplanation},
where we introduce the notion of an explanation $\pi$'s \emph{quality},
taking into account both $\V{\phi, \pi}$ as well as the surprisingness
of $\pi$ itself.
Intuitively we expect that an arbitrary statement $\phi$
ought to have an explanation of sufficiently high quality.
We don't know how to define the quality of an explanation,
but if we use $\abs{\pi}$ as an estimate for the surprisingness
of $\pi$ then we obtain the following stronger form of soundness:
\begin{equation*}
    \exists \delta \colon \text{For all true sentences $\phi$:\,}
    \forall \pi^- \colon \exists \pi^+ \colon
    \abs{\pi^+} - \log \V{\phi, \pi^-, \pi^+} < \abs{\phi} + \abs{\pi^-} + \delta \end{equation*}
The intuitive justification for this principle similar
to the justification for soundness itself but much weaker.
We need to include $\abs{\pi^-}$ based on the concerns
raised in Appendix~\ref{cherrypicking},
and we don't think that this correction term fully handles
the problem raised there.
Nevertheless,
we think it is quite plausible that there is \emph{some} quantitative form
of soundness that is interesting even for finite claims
and is satisfied by an appropriate deductive heuristic estimator.

\section{Cumulant propagation}\label{cumulantappendix}

In Section~\ref{circuitsection} we introduced the problem
of estimating the output probability $\pc$ for a boolean circuit $C$.
In this section we will describe an algorithm for an even simpler problem:
estimating the expected output $\ec$ for an \emph{arithmetic} circuit
$\sig{C}{\mathbb{R}^n}{\mathbb{R}}$
when run on independent Gaussian inputs.
In this simple setting we can improve
over the naive algorithm which simply treats all gates as independent,
by tracking the expectation of every degree $k$ polynomial
for some constant $k$.
We present this algorithm in Section~\ref{cumulantpropalg}.

Often many of these correlations will be small,
and so we'd like to design a faster algorithm
that pays attention to a specific subset of polynomials specified
in an argument $\pi$,
and continues to treat other variables as independent.
Unfortunately,
when we do this our algorithm can produce
inconsistent estimates with $\E{f^2} < 0$ for a real-valued polynomial $f$.
We explore this difficulty in Section~\ref{cumulantproblems}.

We believe that there likely exists
an estimation algorithm
that corrects these deficiencies.
Finding such an algorithm is our current research priority
for formalizing the presumption of independence.

\subsection{Arithmetic circuits}

An arithmetic circuit is exactly analogous to boolean circuits
as defined in Section~\ref{circuitdefinition}
except with node values being real instead of boolean,
additional ``constant wires'' whose value is a fixed constant $c \in \mathbb{R}$,
and operations being either addition or multiplication
rather than an arbitrary boolean function.

Formally,
an arithmetic circuit $C$ with $n$ inputs
consists of a set of nodes $x_1, x_2, \ldots, x_m$.
Each node $x_k$ is labeled as one of:
\begin{itemize}
    \item An input wire labeled with an integer $i_k \in \set{1, 2, \ldots, m}$. 
        For convenience we will assume that there is exactly one input wire with each label.
    \item A constant wire labeled with real number $c_k \in \mathbb{R}$.
    \item A sum gate labeled with a pair $a_k, b_k \in \set{1, 2, \ldots, k-1}$.
    \item A product gate labeled with a pair $a_k, b_k \in \set{1, 2, \ldots, k-1}$.
\end{itemize}
To evaluate $C\of{z_1, \ldots, z_n}$
we iterate through the wires in order and compute a value $x_k\of{z_1, \ldots, z_n}$ for each of them:
\begin{itemize}
    \item If $x_k$ is an input wire labeled with $i$, then $x_k\of{z_1, \ldots, z_n} = z_i$.
    \item If $x_k$ is a constant gate labeled with $c_k$, then $x_k\of{z_1, \ldots, z_n} = c_k$.
    \item If $x_k$ is a sum gate labeled with $a_k, b_k$, then
        $x_k\of{z_1, \ldots, z_n} = x_{a_k}\of{z_1, \ldots, z_n} + x_{b_k}\of{z_1, \ldots, z_n}$.
    \item If $x_k$ is a product gate labeled with $a_k, b_k$, then
        $x_k\of{z_1, \ldots, z_n} = x_{a_k}\of{z_1, \ldots, z_n} \times x_{b_k}\of{z_1, \ldots, z_n}$.
\end{itemize}
Then we define $C\of{z_1, \ldots, z_n} = x_{m}\of{z_1, \ldots, z_n}$.

Given an arithmetic circuit $C$ and a distribution $\dist$ over $\mathbb{R}^m$,
we define $\Eof{\dist}{C} = \Eof{z_1, \ldots, z_n \sim \dist}{C\of{z_1, \ldots, z_n}}$.

We will consider heuristic verifiers for estimating $\Eof{\mathcal{N}\of{0, I}}{C}$,
the expected output of $C$ when run on independent standard normal inputs.
We will later see how to generalize this algorithm to other input distributions.

\subsection{Mean propagation}

If we ignore \emph{all} correlations between intermediate values $x_k$,
we obtain a very simple estimator $\tE$ we call ``mean propagation.''
This is precisely analogous to the simple estimator
for boolean circuits introduced in Section~\ref{meanprop}.

\begin{figure}
    \centering
    \begin{tikzpicture}[
        roundnode/.style={
            circle, draw=black!60, fill=green!5, very thick, minimum size=7mm
        },
        naked/.style={
        },
        ]
        \node[naked] (z1) [label={[text=red]15:$0$}] {$z_1$};
        \node[naked] (one) [left=of z1, label={[text=red]15:$1$}] {$1$};
        \node[naked] (z2) [right=of z1, label={[text=red]15:$0$}] {$z_2$};
        \node[roundnode] (onez1) [below=of one, label={[text=red]15:$1$}] {$+$};
        \node[roundnode] (onez2) [below=of z1, label={[text=red]15:$1$}] {$+$};
        \node[roundnode] (z1z2) [below=of z2, label={[text=red]15:$0$}] {$+$};
        \node[roundnode] (A) [below left=of onez2, label={[text=red]15:$1$}] {$\times$};
        \node[roundnode] (B) [below right=of onez2, label={[text=red]15:$0$}] {$\times$};
        \node[roundnode] (C) [below right=of A, label={[text=red]15:$1$}] {$+$};
        \draw[->] (one) -- (onez1);
        \draw[->] (z1) -- (onez1);
        \draw[->] (z1) -- (z1z2);
        \draw[->] (z2) -- (z1z2);
        \draw[->] (z2) -- (onez2);
        \draw[->] (one) -- (onez2);
        \draw[->] (onez1) -- (A);
        \draw[->] (onez2) -- (A);
        \draw[->] (onez2) -- (B);
        \draw[->] (z1z2) -- (B);
        \draw[->] (B) -- (C);
        \draw[->] (A) -- (C);
    \end{tikzpicture}
    \caption{The result of mean propagation on a simple circuit.
        The values $\mu_i$ are written in red beside
        the corresponding $x_i$.}\label{meanprop-example}
\end{figure}
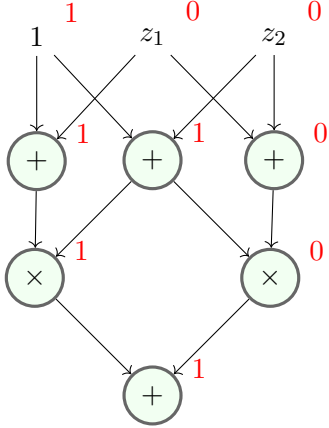

We will iterate through the nodes in order,
and for each node $x_k$ we will compute an estimate $\mu_k$ of its mean as follows:
\begin{itemize}
    \item If $x_k = c_k$ is a constant wire, then $\mu_k = c_k$.
    \item If $x_k = z_{i_k}$ is an input wire, then $\mu_k = 0$
        (since $z_{i_k} \sim N(0, 1)$).
    \item If $x_k = x_{a_k} + x_{b_k}$ is a sum gate,
        then $\mu_k = \mu_{a_k} + \mu_{b_k}$.
        If the estimates for the input wires are accurate, then $\mu_k$ is exactly accurate
        by linearity of expectation.
    \item If $x_k = x_{a_k} x_{b_k}$ is a product gate,
        then $\mu_k = \mu_{a_k} \mu_{b_k}$.
\end{itemize}
An example of this process is illustrated in Figure~\ref{meanprop-example}.

This estimate is ``better than nothing'' in that we expect it to typically
do better than simply assuming the output of a circuit is $0$.
It's not easy to define a formal sense in which we can prove
that this is actually better than nothing,
and so for now we will mostly leave this as an intuitive statement.

Regardless of whether this is better
than nothing, it is certainly not a \emph{great} estimate.
For example, it approximates the mean of $z_i^2$ as $0$ even though $z_i$
is a standard normal.

\subsection{Covariance propagation}

Instead of merely maintaining an estimate for the means of nodes $\mu_k$,
we can also maintain estimates $\sigma_{kk}$ for the variance of each node 
and $\sigma_{jk}$ for the covariance for each pair of nodes.
Now when we consider a new node $k$,
we compute estimates $\sigma_{jk}$ for every node $j \leq k$.
We can compute these estimates using update rules
similar to the last section, but using a slightly more complex ``independence'' assumption.

In particular,
when we are given a product gate like $x_k = x_i x_j$,
we can exactly compute the mean of $x_k$ as $\mu_i \mu_j + \sigma_{ij}$.
But in order to compute the covariance of $x_k$ with $x_{ell}$,
we need to reason about the three-way interaction of $x_i$, $x_j$, and $x_{\ell}$
given only the covariances.
To do this, we will assume that $x_i$, $x_j$, and $x_{\ell}$ are jointly Gaussian.
In this case it is easy to compute that
\[\sigma_{k\ell} = \sigma_{i\ell} \mu_j + \mu_i \sigma_{j \ell}\]
while 
\[\sigma_{kk} = \sigma_{ij}^2 + 2\sigma_{ij}\mu_i\mu_j + \sigma_{ii} \sigma_{jj}
+ \sigma_{ii} \mu_j^2 + \sigma_{jj} \mu_i^2.\]


Why assume that the $x_i$ are jointly Gaussian?
The simplest justification is that this is the maximum entropy distribution
given a particular covariance matrix.
Another intuition is that if they deviate from joint normality,
it's not at all clear which way we should expect the deviation to push.
In Section~\ref{cumulants}
we will generalize this assumption further and give an additional argument
that it is a natural generalization
of the presumption of independence.

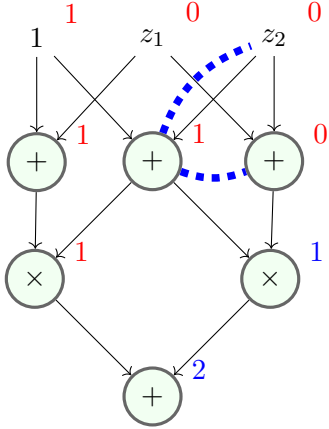
\begin{figure}
    \centering
    \begin{tikzpicture}[
        roundnode/.style={
            circle, draw=black!60, fill=green!5, very thick, minimum size=7mm
        },
        naked/.style={
        },
        ]
        \node[naked] (z1) [label={[text=red]15:$0$}] {$z_1$};
        \node[naked] (one) [left=of z1, label={[text=red]15:$1$}] {$1$};
        \node[naked] (z2) [right=of z1, label={[text=red]15:$0$}] {$z_2$};
        \node[roundnode] (onez1) [below=of one, label={[text=red]15:$1$}] {$+$};
        \node[roundnode] (onez2) [below=of z1, label={[text=red]15:$1$}] {$+$};
        \node[roundnode] (z1z2) [below=of z2, label={[text=red]15:$0$}] {$+$};
        \node[roundnode] (A) [below left=of onez2, label={[text=red]15:$1$}] {$\times$};
        \node[roundnode] (B) [below right=of onez2, label={[text=blue]15:$1$}] {$\times$};
        \node[roundnode] (C) [below right=of A, label={[text=blue]15:$2$}] {$+$};
        \draw[->] (one) -- (onez1);
        \draw[->] (z1) -- (onez1);
        \draw[->] (z1) -- (z1z2);
        \draw[->] (z2) -- (z1z2);
        \draw[->] (z2) -- (onez2);
        \draw[->] (one) -- (onez2);
        \draw[->] (onez1) -- (A);
        \draw[->] (onez2) -- (A);
        \draw[->] (onez2) -- (B);
        \draw[->] (z1z2) -- (B);
        \draw[->] (B) -- (C);
        \draw[->] (A) -- (C);
        \draw[blue, dashed, line width=1mm] (onez2) to[out=-20,in=200] (z1z2);
        \draw[blue, dashed, line width=1mm] (onez2) to[out=70,in=200] (z2) ;
    \end{tikzpicture}
    \caption{In blue we show how covariance propagation gets
        a different estimate than mean propagation.
        The dotted blue lines indicate pair of nodes
        that are computed to have covariance $1$,
        reflecting the computation:
        $1 = \Cov{z_2, z_2} \rightarrow \Cov{1 + z_2, z_2} \rightarrow \Cov{1 + z_2, z_1 + z_2} \rightarrow \E{(z_1 + z_2)(1 + z_2)}$.
        The other covariances are either $0$ or irrelevant to the computation.
    }\label{covprop-example}
\end{figure}

Putting this altogether, the algorithm is:
\begin{itemize}
    \item If $x_k = c_k$ is a constant wire,
        then $\mu_k = c_k$, $\sigma_{jk} = 0$ for all $j$.
    \item If $x_k$ is an input wire, then $\mu_k = 0$, $\sigma_{jk} = 0$ for $j < k$,
        and $\sigma_{kk} = 1$.
    \item If $x_k = x_{a_k} + x_{b_k}$ is a sum wire,
        then $\mu_k = \mu_{a_k} + \mu_{b_k}$,
        $\sigma_{jk} = \sigma_{ja_k} + \sigma_{jb_k}$ for $j < k$,
        and $\sigma_{kk} = \sigma_{a_k a_k} + 2 \sigma_{a_k b_k} + \sigma_{b_k b_k}$.
    \item If $x_k = x_{a_k} x_{b_k}$,
        then 
        \begin{align*}
            \mu_k &= \mu_{a_k} \mu_{b_k} + \sigma_{a_k b_k},\\
            \sigma_{jk} &= \sigma_{ja_k} \mu_{b_k} + \sigma_{jb_k}\mu_{a_k},\\
            \sigma_{kk} &= \sigma_{a_kb_k}^2 + 2\sigma_{a_kb_ke}\mu_{a_k}\mu_{b_k} + \sigma_{a_ka_k} \sigma_{b_kb_k}
            + \sigma_{a_ka_k} \mu_{b_k}^2 + \sigma_{b_kb_k} \mu_{a_k}^2.
        \end{align*}
\end{itemize}
In Figure~\ref{covprop-example} we illustrate how this process
produces a different estimate for the circuit from Figure~\ref{meanprop-example}.

Empirically
we've found this estimate is often much more reasonable than simply propagating means;
for example we've evaluated it for small random circuits or shallow neural networks.
But it remains hard to justify that statement in any formal sense,
or even to justify the claim that it is a ``sound'' estimator,
since it is easy to construct circuits where it gives a worse estimate than nothing.
We could try to formalize soundness by considering particular random distributions
over circuits,
but the results are then very specific to the particular distribution
and do not obviously apply to any realistic circuit.
For now we will mostly leave this as an intuitive claim.

\subsection{Sparse covariance propagation}\label{sparsecovariance}

Covariance propagation gives us a $n^2$ time algorithm
for estimating the output of a circuit with $n$ gates.
If we want a faster algorithm,
we could try to pay attention to a subset of ``important'' covariances.
This introduces a role for arguments $\pi$,
which can point out a set of covariances to pay attention to.


We will take an ``argument'' $\pi$ to be a set of pairs of indices $\set{(i, j)}$
for which we should track covariances.
We compute our estimate exactly as in covariance propagation,
but we only compute covariances $\sigma_{ij}$
for pairs $\of{i, j} \in \pi$.
Whenever a term $\sigma_{ij}$ with $\of{i, j}\not\in\pi$
occurs inside an update step, we replace it with $0$.

Given a set of arguments $\pi_1, \ldots, \pi_n$,
we just apply the same algorithm to the union $\pi = \pi_1 \cup \ldots \cup \pi_n$.

The estimate $V\of{\Eof{\dist}{C}, \pi_1, \ldots, \pi_n}$
can be computed in time $O\of{\abs{C} + \abs{\bigcup_i \pi_i}}$ and
converges to the output of variance propagation
as $\bigcup_i \pi_i \rightarrow C \times C$.
How quickly it converges depends on the details of the circuit
and on how well the arguments $\pi_i$ capture
the important sources of variances.

\subsection{Generalizing independence with cumulants}\label{cumulants}

Mean propagation is organized around the ``naive guess''
$\E{x_i x_j} = \E{x_i} \E{x_j}$,
which we justified by appealing to the presumption of independence.
Covariance propagation is organized
around a similar naive guess, that
$\Cov{x_i x_j, x_{\ell}} = \Cov{x_i, x_{\ell}}\E{x_j} + \Cov{x_j, x_{\ell}}\E{x_i}$,
which we justified by assuming that the $x_i$ were jointly normal
(or equivalently taking a maximum entropy distribution).

In order to deal with higher-order correlations,
we need to generalize these guesses.
We will do this by generalizing a particular definition of independence
based on \emph{joint cumulants}.

For any random variables $X_1, \ldots, X_n$,
the joint cumulants $\k{X_1, \ldots, X_n}$ are defined via the following
identity relating them to the moments:
\begin{equation}\label{cumulantequation}
    \E{X_1 X_2 \ldots X_n} = \sum_{\substack{\text{$\pi$ a partition} \\ \text{of $\set{1, 2, \ldots n}$}}} \prod_{\substack{\set{i_1, i_2, \ldots, i_k} \\ \text{a part of $\pi$}}} \k{X_{i_1}, X_{i_2}, \ldots, X_{i_k}}
\end{equation}
Intuitively,
we often think of the cumulants $\k{X_1, \ldots, X_n}$
as representing the ``intrinsically $\nth$ order'' part
of the expectation $\E{X_1 \ldots X_n}$,
and then we obtain
the full expectation by summing up over contributions
from all of these intrinsic relationships amongst subsets of the variables.

Formally,
many of the nice properties of the joint cumulants
come from an equivalent definition as the coefficients
of the \emph{logarithm} of the moment generating function.
They are also essentially the unique statistic
such that if the $X_i$ are independent from all of the $Y_i$,
then
$\k{X_1 + Y_1, \ldots, X_n + Y_n} = \k{X_1, \ldots, X_n} + \k{Y_1, \ldots, Y_n}$,
and have many other nice properties that lead them to occur
frequently in statistics.

As special cases we have $\k{X} = \E{X}$
and $\k{X, Y} = \Cov{X, Y}$.
We can obtain a recursive definition for $\k{X_1, \ldots, X_n}$
in general by solving Equation~\ref{cumulantequation}.
For example,
\begin{equation}\label{thirdcumulant}
    \k{X, Y, Z} = \E{XYZ} - \E{X}\E{Y}\E{Z} - \Cov{X, Y}\E{Z} - \Cov{X, Z}\E{Y} - \Cov{Y, Z}\E{X}
\end{equation}

If two variables $X$ and $Y$ are independent
then \emph{any} cumulant involving both $X$ and $Y$ (and no other variables)
is zero.
In fact,
for bounded variables this is equivalent to independence.
This suggests a generalization of independence:
we say that $X_1, X_2, \ldots, X_n$ have ``no $n$-way interactions''
if any cumulant involving all of $X_1, \ldots, X_n$ (and no other variables) is zero.
Of course this assumption can be overturned
by noticing an $n$-way interaction,
but we propose it as a reasonable default guess.

This directly allows us to make a guess about $\E{X_1 X_2 \ldots X_n}$
given only lower-order information.
For example, if we know the covariances of $X, Y, Z$
and assume that $\k{X, Y, Z} = 0$,
then Equation~\ref{thirdcumulant}
implies a guess about $\E{XYZ}$.
In fact Gaussians have third and higher cumulants equal to zero,
and so treating $3$ variables as jointly
normal corresponds exactly to this special case with $n = 3$.

We won't try to argue that this is the ``right'' guess,
because we think that it isn't (we'll return to this issue
in Section~\ref{cumulantproblems}).
We do think it is better than nothing and it's not \emph{obvious}
how to improve on it. For example,
if we want to infer $\E{XYZW}$
from knowledge of the second and third moments,
we believe that this algorithm is much better
than simply ignoring the third moments
and treating $X, Y, Z, W$ as jointly Gaussian.
We won't make this claim precise.

Before explaining why we don't yet
think this is the ``right'' answer,
we'll show how we can use
the ``joint cumulants are zero''
assumption in order to write down a
natural generalization of covariance propagation
to handle higher order interactions.

\subsection{Cumulant propagation}\label{cumulantpropalg}

Using cumulants, we can generalize covariance propagation:
instead of estimating the covariances $\Cov{x_i, x_j}$
we estimate the $\nth$ cumulants $\k{x_{i_1}, x_{i_2}, \ldots, x_{i_n}}$.
The update rules are now more complex,
but they can still be derived directly from Equation~\ref{cumulantequation}.
As in covariance propagation,
we consider the nodes in order $x_1, x_2, \ldots x_{m}$.
Whenever we consider a new node $x_k$,
we can estimate the cumulants involving $x_k$
by using the definition of $x_k$,
Equation~\ref{cumulantequation},
and the assumption that unknown cumulants are zero.

In practice,
we find that tracking these higher cumulants continues to improve
our estimates at the expense of additional compute
(we don't report experiments here).

As in sparse covariance propagation,
we can potentially make this algorithm faster
by considering a set $\pi$
of tuples $\of{x_{i_1}, \ldots, x_{i_r}}$
and only tracking cumulants for tuples in $\pi$
(rather than tracking all $\nth$ cumulants for some fixed $n$).
Whenever a cumulant we are not tracking appears in an equation,
we assume that it is zero.
For covariance propagation this only reduced the computational cost from $\abs{C}^2$ to $\abs{C}$,
but if we are considering very large cumulants then this can mean the difference
between exponential and polynomial time.

We present the pseudocode for this procedure in Algorithm~\ref{cumulantalg}.
As written it involves an exponentially large sum over all
partitions of $\set{2, 3, \ldots, r}$,
but the overall algorithm can easily be sped up to $\widetilde{O}\of{\abs{\pi}^2}$
by doing some elementary combinatorics and only considering
non-zero terms in the sum.
\begin{algorithm}
    \caption{Cumulant propagation}\label{cumulantalg}
    \SetKwInOut{Input}{Input}
    \SetKwFunction{FLookup}{cumulant}
    \SetKwProg{Fn}{Function}{:}{}
    \Input{A circuit $C = \of{x_1, x_2, \ldots, x_{m}}$
        and a set $\pi \subset \of{x_1, x_2, \ldots, x_{m}}^*$
    of sequences of variables.}
    Sort each $\of{x_{i_1}, \ldots, x_{i_r}} \in \pi$ so that $i_1 \geq \ldots \geq i_r$\;
    $\of{S_1, S_2, \ldots, S_N} \gets \text{the list of sorted tuples in $\pi$,
    sorted in lexicographic order}$\;
    \For{$i = 1, 2, \ldots, N$}{
        $\kappa\of{S_i} \gets 0$
    }
    \Fn{\FLookup{$x_{i_1}, x_{i_2}, \ldots, x_{i_r}$}}{
        \For{$i = 1, 2, \ldots, N$} {
            \If{$S_i = \set{x_{i_1}, x_{i_2}, \ldots, x_{i_r}}$} {
                \KwRet $\kappa\of{S_i}$
            }
        }
        \KwRet $0$\;
    }
    \For{$i = 1, 2, \ldots, N$}
    {
        $\of{x_{i_1}, \ldots, x_{i_r}} \gets S_i$\;
        \uIf{$x_{i_1} = c$ is a constant gate} {
            \If{$N = 1$} {
                $\kappa\of{S_i} \gets c$\;
            }
        } \uElseIf{$x_{i_1} = z_j$ is an input gate} {
            \If{$N = 2$ and $x_{i_2} = z_j$} {
                $\kappa\of{S_i} \gets 1$\;
            }
        } \uElseIf{$x_{i_1} = x_a + x_b$ is a sum gate} {
            $\kappa\of{S_i} \gets \FLookup\of{x_a, x_{i_2}, \ldots, x_{i_r}}
            + \FLookup\of{x_b, x_{i_2}, \ldots, x_{i_r}}$
        } \ElseIf{$x_{i_1} = x_a * x_b$ is a product gate} {
            $\k{S_i} \gets \FLookup\of{x_a, x_b, x_{i_2}, \ldots, x_{i_r}}$\;
            \For{each partition $\of{
                    \set{j_1, j_2, \ldots, j_a},
                    \set{k_1, k_2, \ldots, k_b}
                }$ of $\set{2, \ldots, r}$} {
                $\k{S_i} \gets  \k{S_i} + 
                    \FLookup\of{x_a, x_{i_{j_1}}, \ldots, x_{i_{j_a}}}
                    \FLookup\of{x_a, x_{i_{k_1}}, \ldots, x_{i_{k_b}}}
                $
            }
        }
    }
    \KwRet $\FLookup\of{x_{m}}$
\end{algorithm}

We have not argued that this is a particularly expressive proof system
for realistic problems.
We offer it primarily to give a concrete illustration
of what a heuristic argument can look like
and how the ``presumption of independence''
can be used to produce anytime estimates for quantities that are very hard to estimate exactly.
We are tentatively optimistic that similar ideas can be generalized to obtain much better estimates
for a broad range of claims,
but for now that is only a vague intuitive hope.
In order to actually obtain good estimates,
we would likely need to address
the many limitations in Algorithm~\ref{cumulantalg}.
In the next section we list some of these problems.

\subsection{Cumulant propagation and sums of squares}\label{cumulantproblems}

Cumulant propagation
satisfies all of the desiderata in Section~\ref{desiderata},
except for \emph{respect for proofs}.
In particular,
cumulant propagation often produces negative estimates $\V{X^2} < 0$
even though it is easy to prove that $X^2 \geq 0$.

We could fix this problem by truncating cumulant propagation's estimates---whenever
we have $\V{f} < 0$,
and $\pi$ is a representation of $f$ as a sum of squares,
then we could define $\V{f, \pi} = 0$.
We consider this response highly unsatisfying. In addition to throwing
away all the information that went into the estimate of $\V{f}$,
and producing an implausible estimate exactly on the boundary of possibility,
a simple version of this approach will also violate linearity of expectation.

Another approach would be to simply ignore
the arguments raised by cumulant propagation.
For example, if $\E{A} = \E{B} = \E{C} = \E{D} = 0$,
but $\Cov{A, B} = \Cov{C, D} = 1$
then we could treat $\E{ABCD} = \E{A}\E{B}\E{C}\E{D} = 0$ by default.
We consider this unsatisfying because
the non-zero covariances feel like
a strong \emph{prima facie} argument
about the value of $\E{ABCD}$.
Neglecting it seems to be ignoring important information,
and once we go down that road it seems plausible
we need to neglect essentially \emph{all} information.
This can be true even if
this type of argument,
taken in isolation,
can lead to a clearly
unreasonable estimates.

Instead we'd like to find a heuristic estimator that allows
us to capture the kinds of considerations raised by cumulant propagation
while respecting the coherence properties in Section~\ref{desiderata}---including
respect for sum-of-squares proofs.

Although sum-of-squares is a specific proof system,
it is quite powerful and flexible\footnote{
    For example, positive expectations for sums of squares
    is a sufficient condition for a set of moments to be realizable,
    and sum of squares proofs play a central role in the
    theory of approximate constraint satisfaction.
},
and we think that finding an estimator
that respects sum-of-squares proofs would be a major
step towards formalizing the presumption of independence in general.
Although we won't discuss the connection in detail,
cumulant propagation also assigns estimates $\V{C} \not\in[0, 1]$
for arithmetizations of boolean circuits run on boolean inputs,
and we believe this failure is closely connected to negativity of squares.


%
In this section we will briefly discuss a few cases where cumulant
propagation can produce a negative estimate for an expectation $\V{X^2}$.

\subsubsection{Imputing missing moments}

Suppose that I'm tracking the following means and covariances:
\begin{gather*}
    \E{X} = \E{Y} = \E{Z} = 0, \\
    \Var{X} = \Var{Y} = \Var{Z} = 1, \\
    \Cov{X, Y} = \Cov{Y, Z} = 0.9, \\
\end{gather*}
but I don't know $\Cov{X, Z}$.
That is, my beliefs about the covariance of $X$, $Y$, $Z$ are represented by the matrix
\[ \begin{pmatrix}
    1 & 0.9 & ?? \\
    0.9 & 1 & 0.9 \\
    ?? & 0.9 & 1
\end{pmatrix} \]

Now suppose we calculate $\E{(X - 2Y + Z)^2}$ by filling in the missing entry as zero:
\begin{align*}
    \E{(X - 2Y + Z)^2} &= \E{X^2} + 4\E{Y^2} + \E{Z^2} - 4 \E{XY} - 4 \E{YZ}  + 2 \E{XZ} \\
                       &= 6 - 4 \Cov{X, Y} - 4 \Cov{Y, Z} + 2 \Cov{X, Z} \\
                       &= -1.2 + 2 \Cov{X, Z} \\
                       &< 0
\end{align*}
Cumulant propagation assumes that $\Cov{X, Z} = 0$ since it is unknown,
and therefore estimates $\E{(X-2Y+Z)^2} < 0$.

If we apply maximum entropy with the known covariances,
we would instead make the default guess 
\[\Cov{X, Z} = \frac{\Cov{X, Y} \Cov{Y, Z}}{\Var{Y}}\]
and this guess would guarantee that $\E{p(X, Y, Z)^2} \geq 0$
for any polynomial $p$.
Other perspectives suggest the same heuristic estimate in this case,
and we think it it's quite likely to be the ``right'' one.

We could fix this problem in $O\of{n^2}$
time by finding the maximum entropy distribution consistent
with a given set of moments,
or by simply ensuring that we track \emph{every} second moment.
So we only produce negative estimates
if we try to use sparsity to do a heuristic evaluation
in time $o\of{n^2}$.

We consider this a problem,
but we expect some readers to be less concerned since fixing
the problem requires only a polynomial time slowdown.
Unfortunately,
the same problem can occur when imputing higher moments,
and in that case we do not know how to fix it without an exponential slowdown.

As a simple example, suppose that we are tracking the following cumulants:
\begin{align*}
    \E{X} = 0,\\
    \Var{X} = 1, \\
    \kappa_3\of{X} = \k{X, X, X} = 0, \\
    \kappa_4\of{X} = 100, \\
    \kappa_5\of{X} = 0,
\end{align*}
but are not tracking $\kappa_6\of{X}$.
Then cumulant propagation assumes it is zero,
and so $\E{X^6} = 15 \kappa_2\of{X}^3 + 15 \kappa_2\of{X} \kappa_4\of{X} = 1515$.
But that implies:
\begin{align*}
    \E{\of{100 X - X^3}^2}
    &= 10000\E{X^2} - 200 \E{X^4} + \E{X^6} \\
    &= 10000 - 20000 + 1515 \\
    &= -8485 < 0
\end{align*}
This suggests that assuming $\kappa_6\of{X} = 0$
is not reasonable.

In fact there is no maximum entropy distribution
subject to these limitations.
You can obtain entropy arbitrarily close to a Gaussian with variance $1$
by taking $X$ to be a mixture with $(1 - \epsilon)$ probability
of being Gaussian and $\epsilon$ probability of being equal to
$\sqrt[4]{\frac {97}{\epsilon}}$.
In the limit as $\epsilon \rightarrow 0$ the entropy approaches
the entropy of a Gaussian,
while $\kappa_6\of{X} \rightarrow \infty$.
Regardless of whether or not $\kappa_6\of{X} = \infty$
is a reasonable best guess,
this differs considerably from cumulant propagation
and we don't think it can serve as the basis
for a reasonable algorithm for heuristic evaluation of arithmetic circuits.

A similar failure can arise if we know the joint distribution of
any set of $5$ variables from $X_1, X_2, \ldots, X_6$
but don't know the cumulant $\k{X_1, X_2, \ldots, X_6}$.
In this case there is a maximum entropy
distribution,
for which $\E{X_1 X_2 \ldots X_6}$ can be expressed
as a sum of rational functions of the known moments of the $X_i$.
But we do not know how to approximate this best guess polynomial time
and so we are interested in computationally tractable approximations
which still respect coherence properties.

\subsubsection{Sparse covariance propagation for linear circuits}

So far we've talked about inferring missing moments
and argued that the way cumulant propagation handles this problem
will lead directly to negative expectations for squares.
But it's not clear that a successful alternative
to cumulant propagation needs to ever infer missing moments,
rather than following a completely different strategy.
In this section and the next one,
we describe estimation problems where we don't know any reasonable efficient
heuristic estimator.

Suppose that $z_1, \ldots, z_n$ are independent Gaussian inputs,
that $A \in \mathbb{R}^{m \times m}, B \in \mathbb{R}^{m \times m}$ are linear maps,
and that $y = A z$ is a vector of $m$ intermediates,
and $x = B y$ is a vector of $m$ outputs.
Suppose that we want to estimate each of the variances $\Var{x_i}$.

We can compute exactly that 
\begin{align*}
    \Var{x_i} &= \Cov{\sum_j B_{ij} y_j, \sum_j B_{ij}, y_j} \\
              &= \sum_{j,k} B_{ij} B_{ik} \Cov{y_j, y_k} \\
              &= \sum_{j,k} B_{ij} B_{ik} \Cov{\sum_l A_{jl} z_l, \sum_l A_{kl} z_l} \\
              &= \sum_{j,k, l} B_{ij} B_{ik} A_{jl} A_{kl}
\end{align*}
So estimating the variances of the $x_i$
amounts to computing all $n$ of these sums.
Each sum involves $m^3$ terms.
We can compute all of these sums at once by doing 3 matrix multiplications,
in time $O\of{m^{\omega}}$,
but we are interested in finding a significantly faster algorithm.

An equivalent way to think about this sum is that we are given a list of vectors $v_i$
corresponding to the rows of $A$,
and we want to compute $\lvert \sum \alpha_i v_i \rvert^2$
for $m$ different vectors $\alpha$.

Sparse covariance propagation corresponds to one way to estimate this sum.
By tracking only $O\of{m}$ of the terms $\Cov{y_j, y_k}$,
we can obtain the following estimate in time $O\of{m^2}$:
\begin{itemize}
    \item Choose a list $S$ of $O\of{m}$ pairs $(j, k)$.
    \item For each pair $(j, k) \in S$, compute $\cov{y_j, y_k} = \of{AA^T}_{jk} = \sum_l A_{jl} A_{kl}$.
        Each of these $O\of{m}$ sums takes $m$ time to compute.
    \item For each $i$, compute $\sum_{(j, k) \in S} B_{ij} B_{ik} \of{AA^T}_{jk}$.
        Use this as our estimator for $\Var{x_i}$.
        Each of these $m$ sums takes $O\of{m}$ time to compute.
\end{itemize}
We wanted to calculate
$\Var{x_i}$ which is a sum of $m^3$
terms of the form $B_{ij} B_{ik} A_{jl} A_{kl}$.
This approximation takes a sum of $O\of{m^2}$ terms
and then approximates the rest as $0$.

In some cases this sparse approximation captures a significant
part of the full sum.
For example, if each row of $A$ is obtained by applying
a random small rotation to the previous row,
then $\sum_l A_{jl} A_{kl}$ decays exponentially with $\abs{j-k}$,
and so taking the set of terms near the diagonal can give you an extremely good approximation.

However, this estimate can be negative in a way that exactly
mirrors the failure discussed in the previous section,
so it's clearly not the \emph{most} reasonable estimate.
Suppose that we take $S = \set{(j, k) : \abs{j - k}\leq 1}$,
and compute $\Var{y_j} = 1$ and $\Cov{y_j, y_{j+1}} = 0.9$.
The case $n = 5$ is illustrated in Figure~\ref{chain}.

\begin{figure}
    \centering
    \[
    \begin{pmatrix}
        1 & 0.9 & ? & ? & ?\\
        0.9 & 1 & 0.9 & ? & ?\\
        ? & 0.9 & 1 & 0.9 & ?\\
        ? & ? & 0.9 & 1 & 0.9\\
        ? & ? & ? & 0.9 & 1\\
    \end{pmatrix}
    \]
    \caption{Suppose $(AA^T)_{ij} = \Cov{y_i, y_j}$ is indicated above,
        where we've computed covariances only for $\abs{i - j} \leq 1$.
        If we simply drop the terms marked $?$,
        we obtain the estimate $\Var{y_1 - y_2 + y_3 - y_4 + y_5} = -2.2$.
        It would be better to make the maximum
        entropy guess $0.9^k$ for the covariances
        $\Cov{y_i, y_j}$ with $\abs{i - j} = k$.
        This results in the estimate around $+1$ instead of $-2.2$.
        For tree sparsity patterns, it is possible to compute this maximum
    entropy estimate for $\Var{\sum \alpha_i y_i}$ in time $O\of{m}$.} \label{chain}
\end{figure}

Now consider the case where a row of $B$ consists of alternating signs,
i.e where $x_i = \sum_{j} \of{-1}^j y_j$.
Our estimate is:
\begin{align*}
    \Var{x_i} &= \sum_j \Var{y_j} + 2 \sum_j \Cov{y_j, y_{j+1}} \\
              &= m - 1.8 (m - 1) \\
              &< 0
\end{align*}
At this point it's not clear what we should estimate for $\Var{x_i}$.
We were interested in the sum of $m^3$ terms,
which we knew would be positive.
We've added up $m^2$ of those terms and found the sum to be negative.
The question is what estimate we give for the remaining terms.
Simply estimating $\Var{x_i} = 0$ amounts to assuming
that the unobserved terms exactly cancel the observed terms,
which seems like a bad estimate that throws away information.

In the special case where we know $\Var{y_j}$ and $\Cov{y_j, y_{j+1}}$
we believe this question has a nice answer.
Namely,
we should make the maximum entropy assumption that
\[ \Cov{y_j, y_k} = \frac{ \Cov{y_j, y_{j+1}} \Cov{y_{j+1}, y_{j+2}} \ldots \Cov{y_{k-1}, y_k}}
{\Var{y_{j+1}}\Var{y_{j+2}}\ldots\Var{y_{k-1}}}.\]
It turns out that this always results in a non-negative
estimate for $\Var{x_i}$,
and moreover that the estimate can be computed in linear time using dynamic programming.

We don't know whether it is possible to generalize this algorithm.
But at any rate,
we think that it should be possible to find a better estimate than $\Var{x_i} = 0$.
If this is \emph{not} possible
then in our view it calls into question some of our optimistic intuitions
about how anytime estimates should work and why it should be possible to produce them.

\subsubsection{Estimating the permanent of a PSD matrix}

In this section we describe an estimation problem where we don't know
how to obtain reasonable estimates in polynomial time.
We discuss the connection to cumulant propagation at the end of the section.

For an $n \times n$ matrix $A$,
define the permanent
\[ \perm\of{A} = \sum_{\sigma} \prod_i A_{i\sigma\of{i}} \]
where the sum is taken over every permutation $\sig{\sigma}{\range{n}}{\range{n}}$.
Computing or even approximating the permanent is very difficult.

One way to learn about $\perm\of{A}$
is to compute
\[\V{\perm\of{A}, \pi} = \sum_{\sigma \in \pi} \prod_i A_{i\sigma\of{i}} \]
for a particular set of permutations $\pi = \set{\sigma_1, \ldots, \sigma_m}$.
As discussed in Section~\ref{sumexample},
computing the sum of a subset of terms gives us a heuristic estimate
for the full sum.
This is usually a poor estimate unless the set $\pi$
is exponentially large.
But if $A$ is very structured or sparse it can be possible
for a small set of terms to capture a significant part of the sum,
and so this heuristic argument can sometimes have a meaningful effect.

If $A$ is positive semi-definite,
i.e. if it can be written in the form $A_{ij} = \inner{v^i}{v^j}$
for a list of vectors $v^i \in \mathbb{R}^n$,
then $\perm\of{A}$ can be written as a sum of squares and so must be non-negative:
\[ \perm\of{A} = \frac 1{n!} \sum_{r_1, r_2, \ldots, r_n \in \range{n}}
\of{\sum_{\sigma} \prod_i v^{\sigma\of{i}}_{r_i}}^2.\]
The exact form of this sum of squares is not important;
what matters is that we have a simple proof of non-negativity.\footnote{
    No algorithm is known for computing the permanent
    even for PSD matrices.
    The best known approximation is given by
    \cite{permanentestimation} and has exponential error.}

Unfortunately,
we can have $\V{\perm\of{A}, \pi} < 0$.
This leaves us in the same situation as in the preceding two sections:
clearly we'd be better off just outputting $0$ rather than
a negative estimate for $\perm\of{A}$.
But outputting $0$ involves assuming that the unobserved
terms in the sum $\perm\of{A}$ exactly cancel out the observed terms,
which again seems like a bad estimate that throws away information
and leads to incoherence.
So it's natural to ask: can we do better?

We are aware of a strictly better estimator in the special case
where the permutations $\sigma_1, \ldots, \sigma_m$ commute
and therefore generate an abelian group $G \subset S_n$.
In this case it turns out to be possible to construct
a set of random variables
such that each term $\prod_i A_{i \sigma\of{i}}$
is a pairwise correlation.
We can then obtain a reasonable estimate of $\perm\of{A}$
by making a maximum entropy assumption about those variables.
Unfortunately, it is not clear how to generalize
this idea to general sets of permutations $\pi$.

Computing $\perm\of{A}$ for a PSD matrix $A$ is
closely related to computing $\E{\of{X_1 \ldots X_n}^2}$
where the $X_i$ have covariance matrix $A$.
In fact, $\E{\of{X_1 \ldots X_n}^2}$
is represented by the same sum as the permanent,
but where each term $\prod_i A_{i \sigma\of{i}}$
is multiplied by a factor of $2^{\abs{\sigma}}$
where $\abs{\sigma}$ is the number of cycles in $\sigma$.
Pointing out particular non-zero terms
is one way to approximate this sum,
and this corresponds to cumulant propagation
when the set of observed cumulants
takes a particular special form.
Thus cumulant propagation can produce
negative estimates for $\E{\of{X_1 \ldots X_n}^2}$
in a way that is analogous to our negative estimates
for the permanent.
The factor of $2^{\abs{\sigma}}$
means that the two problems aren't exactly equivalent,
but similar difficulties seem to arise in both cases.
Moreover, a reasonable heuristic estimator
should ultimately be able to handle both of these cases,
and so we regard it as a reasonable test case
for formalizing heuristic arguments.

\section{Cherry-picking arguments}\label{cherrypicking}

In Section~\ref{cherrypickshort} we argued that heuristic
arguments don't always bring our estimates closer to reality.
That is, if we form an estimate based on adversarially chosen arguments
then we can reliably do worse than if we had made a completely naive guess.
This is a difference from the situation with proofs,
where a proof always gets you closer to the truth no matter where it came from.

In this section we present a few examples showing that various
simple fixes do not address the problem.
We then discuss why we think heuristic estimators
are valuable despite these limitations,
and suggest a weaker convergence bound that we think may be achievable.

\subsection{Arguments can make estimates worse}

All of our examples will involve quantities of the form $X = \sum_{x \in \X} \alpha_x f\of{x}$.
We will assume that $\V{f(x)} = 0$ for a generic $x$,
i.e. that $\tE$ sees no reason that $f$ should be biased to be positive or negative.
We'll also assume that $\tE$ sees no correlation between different values of $f$,
and more generally that the only way $\tE$ ever changes its mind about any value $f(x)$
is by computing it.

For any $x \in X$,
we write $\pi_x$ for the argument that exactly calculates a single value $f(x)$.
As discussed in Section~\ref{sumexample},
we expect a reasonable heuristic estimator to satisfy:
\[ \V{X, \pi_{x_1}, \ldots, \pi_{x_k}} = \sum \alpha_{x_k} f\of{x_k}.\]

In Section~\ref{cherrypickshort} we considered finite sums $\sum_{x=1}^n f(x)$
where each $f\of{x} = \pm 1$.
We observed that typically there will be particular values $f(x)$
which have the opposite sign from $X$.
For any such $x$, $\V{X, \pi_x}$ will be a worse estimate than $\V{X}$.
If $x_1, \ldots, x_k$ is the list of all $x$ for which $f(x)$ has the opposite sign from $X$,
then $\V{X, \pi_{x_1}, \ldots, \pi_{x_k}}$
can be an arbitrarily bad estimate for $X$.

\subsection[convergence]{$\Vpi{X}$ does not always converge}

Although it is possible to cherry-pick arguments pointing in the wrong direction,
we might still hope that if we give $\tE$ \emph{enough} good arguments
then it will eventually converge to the truth,
and that the resulting
correct estimate will be robust even if we supply additional cherry-picked arguments.

Unfortunately this does not seem to be the case in general.
Suppose that
\[X = \sum_{x = 1}^{\infty} \frac {f(x)}{x^s}\]
for a constant $1/2 < s < 1$,
where each $f(x)$ is $\pm 1$.

Then we have 
\begin{align*}
    \E{X^2} &= \sum_{x = 1}^{\infty} \frac{f(x)^2}{x^{2s}} \\
             &= \sum_{x = 1}^{\infty} \frac 1{x^{2s}} \\
\end{align*}
which converges for every $s > 1/2$.
Thus $\E{X^2}$ is finite and so $X$ is finite almost surely.
(This is a probabilistic argument that we are making on the outside,
not a heuristic argument that $\tE$ is evaluating.)

But on the other hand, we have:
\begin{align*}
    \E{\sum_{x : f(x) > 0} \frac {f(x)}{x^s}}
    &= \sum_{x = 1}^{\infty} \frac {1}{2 x^s} \\
    &= \infty
\end{align*}
As a result,
no matter how many arguments $\pi_x$ we have seen,
it's most likely the case that the estimate $\V{X}$
can be driven arbitrarily high by presenting additional arguments
$\pi_x$ for $x$ with $f(x) > 0$.
Similarly, $\V{X}$ can be driven arbitrarily low
by presenting $\pi_x$ for $f(x) < 0$.

This issue is clearest in the case of infinite sums,
where $\tE$ literally never converges.
However this also corresponds to a serious quantitative failure for finite sums:
even if the variance of $\sum_{x \in \X} \alpha_x f(x)$ is $\epsilon^2$,
cherry-picking arguments can still lead us to overestimate or underestimate
$X$ by $\epsilon \sqrt{\abs{X}}$,
and we do not converge until $\tE$ has computed the value $f(x)$
for a large fraction of all $x \in \X$.

\subsection{Debate does not lead to convergence}

So far we've argued that there \emph{exist} arguments $\pi$
that would cause $\tE$ to produce bad estimates.
But instead of considering adversarially
chosen arguments designed to mislead, we could imagine
the result of a debate where some arguments are chosen to
make $\V{X}$ large and others are chosen to make $\V{X}$ small.
That is, we could consider the estimate
\begin{equation*}
\max_{\pi_1} \min_{\pi_2} \max_{\pi_3} \ldots \Vpi{X},
\end{equation*}
perhaps with a restriction on the length of each argument $\pi_i$.

Unfortunately this approach also does not produce good estimates.
For example suppose that instead of $f(x) = \pm 1$,
each $f(x)$ has a $1/3$ probability of being equal to $2$ and a $2/3$ probability
of being equal to $-1$.
And suppose each argument $\pi_{x}$ has equal length.
Then consider the same sum as before:
\[X = \sum_{x = 1}^{\infty} \frac {f(x)}{x^s},\]
for a constant $1/2 < s < 1$.

It's easy to see that $X$ almost surely converges to a finite value.
But arguments that $X$ is large are ``more efficient''
since each of them gives us a value where $f(x) = 2$,
while each argument that $X$ is small gives us one value where $f(x) = -1$.
This means that in the limit our estimates for $X$ converge to $+\infty$
instead of the correct finite value.

More precisely, let $x^+_1, x^+_2, \ldots$ and $x^-_1, x^-_2, \ldots$
be the enumeration of integers $x$ where $f(x) = 2$ and $f(x) = -1$
respectively.
Then $x^+_k$ is roughly $3k$,
while $x^-_k$ is roughly $(3/2)k$,
so we have:
\begin{align*}
    \max_{\pi_1} \min_{\pi_2} \max_{\pi_3} \ldots \Vpi{X}
    &=\V{X, \pi_{x^+_1}, \pi_{x^-_1}, \ldots, \pi_{x^+_{n/2}}, \pi_{x^-_{n/2}}} \\
    &= \sum_{i = 1}^{n/2} \frac {f\of{x^+_i}}{\of{x^+_i}^s} +
        \sum_{i = 1}^{n/2} \frac {f\of{x^-_i}}{\of{x^-_i}^s} \\
    &= 2 \sum_{i = 1}^{n/2} \frac {1}{\of{x^+_i}^s} - \sum_{i = 1}^{n/2} \frac 1{\of{x^-_i}^s} \\
    &\approx 2 \sum_{i = 1}^{n/2} \frac 1{(3 i)^s} - \sum_{i = 1}^{n/2} \frac 1{\of{3i/2}^s} \\
    &= \of{\frac 2{3^s} - \frac {2^s}{3^s}} \sum_{i = 1}^{n/2} \frac 1{i^s} \\
    &\rightarrow  \infty
\end{align*}

\subsection{Provable bounds do not lead to convergence}

So far we've seen problems for quantities $X$ that are defined
as convergent but not absolutely convergent series,
for which there is no provable bound on $X$.
We might hope that if we can prove $\ell \leq X \leq h$
then we can converge in finite time
and bound the damage done by cherry-picking based on $h - \ell$.

Unfortunately this also seems to be impossible.

Define the function $\sig{\sigma}{\mathbb{R}}{[-1, 1]}$ via
\[ \sigma\of{x} = \begin{cases}
        -1 & \text{for $x < -1$} \\
        \phantom{-}x & \text{for $-1 < x < 1$} \\
        \phantom{-}1 & \text{for $1 < x$}
    \end{cases}
\]
Consider the quantity
\[ X = \inf_{n} \newsup_{N > n} \sigma\of{\sum_{x = 1}^N \frac {f(x)}{x^{2/3}}}, \]
where $f(x) = \pm 1$ is unbiased and independent for different values of $x$.
If we choose any set $x_1 < \ldots < x_k$ such that 
\[\sum_{i=1}^k \frac{f(x_i)}{x_i^{2/3}} < -100,\]
then we claim that $\V{X, \pi_{x_1}, \ldots, \pi_{x_k}} \approx -1$.
This is because $\tE$'s estimate for variance of $\sum \frac{f(x)}{x^{2/3}}$ is about $3.6$,
and so it assigns a $<0.1\%$ chance that the sum of the remaining terms is more than $99$,
and by a more careful analysis and union bound we could compute
that it assigns at most a $<1\%$ chance that any of the partial sums of the remaining terms
is ever more than $99$.
It therefore has less than a $1\%$ chance that any of the partial sums for $N > x_k$
is ever more than $99$,
and hence less than $1\%$ chance that the inf sup is 
more than $-1$.
As a result, $\tE$ should be at most $-0.99$.

Similarly, if we choose a set of $x_i$ for which the partial sum is more than $100$,
we have $\V{X, \pi_{x_1}, \ldots, \pi_{x_k}} > 0.99$.

Because $\sum x^{-2/3} \rightarrow \infty$,
no matter how many $x_i$ we have already calculated,
we can always find a suitable larger set of $x_i$ for which the sum is either less than $-100$
or more than $100$.
As a result $\tE$ never converges but can be made to oscillate back
and forth between $-1$ and $1$ forever, regardless of the true value of $X$.

(By combining this with a variant of the counterexample from the last section,
we can also obtain a case where a debate would oscillate forever.)

\subsection{Where this leaves us}

Heuristic estimates $\Vpi{X}$ can be systematically inaccurate
if the arguments $\pi_1, \ldots, \pi_n$ are adversarially chosen.
They fail to converge even if we have a provable bound on $X$.
And eliciting arguments from two competing debaters does not address this difficulty.

This suggests that we need to be careful when interpreting heuristic estimates
derived from untrusted arguments.
In order to produce robust estimates \emph{conditioned}
on the set of arguments $\pi_1, \ldots, \pi_n$
we would need to have reasonable beliefs about how the arguments $\pi_i$
were selected and then revise our beliefs not only based on the content
of those arguments but also based on the evidence about the process
that produced those arguments.
For example, if we see a particular argument $\pi$
and know that it was chosen to maximize $\V{X, \pi}$,
then we would need to update our beliefs based on the fact that no stronger argument was found.
This kind of reasoning cannot be captured in the setting of a heuristic estimator
that makes no assumptions about how the arguments $\pi_i$ were selected.

We do not think that these issues interfere with interpreting $\tE$
as a reasonable belief in light of the arguments $\pi_1, \ldots, \pi_n$,
in the case where those arguments were not cherry-picked.
Moreover, we think that studying heuristic estimators
can still clarify a key part of how we should revise our beliefs based
on the contents of arguments,
even if it does not capture fully general Bayesian reasoning about the source
of those arguments.

Fortunately,
it currently seems like these issues are restricted
to ``poorly behaved'' functions $X$,
rather than occurring for arbitrary quantities.\footnote{It
    seems plausible that there is some analog of absolute integrability
    which would cause $\tE$ to converge,
    but it is not clear how to define such a notion
    and disappointing that it would not follow from a provable bound.
}
This is what makes it plausible that we can achieve our ambitious goal in Section~\ref{goal},
which effectively requires that $\tE$ quickly converge to a reasonable estimate.
If it turned out that a more subtle version of
cherry-picking could cause convergence problems when estimating arbitrary quantities $X$,
it would make this goal impossible and would call into question
the entire project of formalizing heuristic arguments.

%

\section{Applications to machine learning}\label{alignment}

Our interest in heuristic arguments is ultimately motivated
by potential applications to machine learning.
We'll briefly describe this motivation here,
but mostly defer the discussion to future articles.

In modern machine learning,
we understand the behavior of large neural networks
primarily by running them on a huge number of examples.
To select a model,
we pick parameters that perform well on a set of training examples (``empirical risk minimization'').
To determine that a model is safe, we 
measure its behavior on a set of held out validation examples.

Empirical risk minimization has a hard time estimating low-probability risks,
predicting the behavior of a system on novel input distributions,
or identifying when a model is giving an answer for an unexpected reason.
We are concerned that over the long term these limitations could lead to catastrophic alignment failures.

Researchers in AI alignment are extremely interested in other strategies
for learning about models that could overcome these limitations of empirical risk minimization,
including \emph{interpretability} and \emph{formal verification}.
But in practice both approaches are quite difficult to apply to state of the art models,
and there are plausible stories for why these might be fundamental difficulties:
\begin{itemize}
    \item Interpretability typically aims to help humans
        ``understand what the model is doing.''
        But it's not clear whether all models actually operate
        in a way that is amenable to human understanding,
        or even exactly what we mean by ``understanding.''
    \item Formal verification is an incredibly demanding standard which delivers perfect confidence.
        It's not clear we have any right to expect formal proofs even for very simple
        properties of very small models.
\end{itemize}
We are interested in formalizing heuristic arguments
because they seem like a third option for analyzing ML systems
that might be easier than either interpretability or formal verification.
\begin{center}
    \begin{tabular}{c||c|c}
        & Human understandable & Machine verifiable  \\
        \hline \hline
        Confident and final & & Formal proof\\
        \hline
        Uncertain and defeasible & Interpretability & \textbf{Formal heuristic argument}
    \end{tabular}
\end{center}
More concretely,
we are particularly interested in two applications of formal heuristic arguments:
\begin{description}
    \item[Avoiding catastrophic failures.]
        Heuristic arguments can let us better estimate the probability
        of rare failures,
        or failures which occur only on novel distributions where we cannot easily draw samples.
        This can be used during validation to estimate risk,
        or potentially during training to further reduce risk.
    \item[Eliciting latent knowledge.]
        Heuristic arguments may let us see ``why'' a model makes its predictions.
        We could potentially use them to distinguish cases where similar behaviors
        are produced by very different mechanisms---for example distinguishing
        cases where a model predicts that a smiling human face will show up on camera
        because it predicts there will actually be a smiling human in the room,
        from cases where it makes the same prediction because it predicts
        that the camera will be tampered with.
        Achieving this goal requires a ``deductive'' heuristic estimator
        in the sense described in Section~\ref{explaining}.
\end{description}
Neither of these applications is straightforward,
and it should not be obvious that heuristic arguments would allow
us to achieve either goal.
But we hope they can illustrate the kind of application of heuristic
estimators we have in mind,
and to help explain our optimism that new strategies
for reasoning about learned models could open new angles of attack on AI alignment.
We'll discuss these applications in much more detail in future articles.

\end{document}